\documentclass{article} 
\usepackage{iclr2023_conference,times}

\usepackage{amsmath,amsfonts,bm}

\def\eqref#1{equation~\ref{#1}}

\def\1{\bm{1}}

\DeclareMathAlphabet{\mathsfit}{\encodingdefault}{\sfdefault}{m}{sl}
\SetMathAlphabet{\mathsfit}{bold}{\encodingdefault}{\sfdefault}{bx}{n}

\DeclareMathOperator*{\argmax}{arg\,max}

\usepackage{hyperref}
\usepackage{url}

\usepackage[utf8]{inputenc} 
\usepackage[T1]{fontenc}    
\usepackage{hyperref}       
\usepackage{url}            
\usepackage{booktabs}       
\usepackage{amsfonts}       
\usepackage{nicefrac}       
\usepackage{microtype}      
\usepackage{xcolor}         
\usepackage{natbib}

\usepackage{graphicx}
\usepackage[linesnumbered,lined,ruled]{algorithm2e}
\usepackage{wrapfig}
\usepackage[shortlabels]{enumitem}

\usepackage{xcolor,colortbl}
\definecolor{mygray}{gray}{0.9}

\newcommand{\hsedit}[1]{{\color{black} #1}}

\newcommand{\eref}[1]{(\ref{#1})}

\title{Symbolic Physics Learner: Discovering governing equations via Monte Carlo tree search}

\iclrfinalcopy

\author{Fangzheng Sun \\
Northeastern University \\
Boston, MA, USA \\
\texttt{sun.fa@northeastern.edu} \\
\And
Yang Liu \\
University of Chinese Academy of Sciences \\
Beijing, China \\
\texttt{liuyang22@ucas.ac.cn} \\
\And
Jian-Xun Wang \\
University of Notre Dame \\
Notre Dame, IN, USA \\
\texttt{jwang33@nd.edu} \\
\And
Hao Sun\thanks{Corresponding author} \\
Renmin University of China \\
Beijing, China \\
\texttt{haosun@ruc.edu.cn}
}

\begin{document}

\maketitle

\begin{abstract}
    Nonlinear dynamics is ubiquitous in nature and commonly seen in various science and engineering disciplines. Distilling analytical expressions that govern nonlinear dynamics from limited data remains vital but challenging. To tackle this fundamental issue, we propose a novel Symbolic Physics Learner (SPL) machine to discover the mathematical structure of nonlinear dynamics. The key concept is to interpret mathematical operations and system state variables by computational rules and symbols, establish symbolic reasoning of mathematical formulas via expression trees, and employ a Monte Carlo tree search (MCTS) agent to explore optimal expression trees based on measurement data. The MCTS agent obtains an optimistic selection policy through the traversal of expression trees, featuring the one that maps to the arithmetic expression of underlying physics. Salient features of the proposed framework include search flexibility and enforcement of parsimony for discovered equations. The efficacy and superiority of the \hsedit{SPL} machine are demonstrated by numerical examples, compared with state-of-the-art baselines.
\end{abstract}

\section{Introduction}

We usually learn the behavior of a nonlinear dynamical system through its nonlinear governing differential equations. These equations can be formulated as $\dot{\mathbf{y}}(t) = d\mathbf{y}/dt = \mathcal{F}(\mathbf{y}(t))$, where $\mathbf{y}(t) = \{y_1(t), y_2(t), ..., y_n(t)\}\in \mathbb{R}^{1\times n_s}$ denotes the system state at time $t$, $\mathcal{F}(\cdot)$ a nonlinear function set defining the state motions and $n_s$ the system dimension. The explicit form of $\mathcal{F}(\cdot)$ for some nonlinear dynamics remains underexplored. For example, in a mounted double pendulum system, the mathematical description of the underlying physics might be unclear due to unknown viscous and frictional damping forms. These uncertainties yield critical demands for the discovery of nonlinear dynamics given observational data. Nevertheless, distilling the analytical form of governing equations from limited noisy data, commonly seen in practice, is an intractable challenge.

Ever since the early work on the data-driven discovery of nonlinear dynamics \citep{dvzeroski1993discovering,dzeroski1995discovering}, many scientists have stepped into this field of study. \hsedit{During} the recent decade, the escalating advances in machine learning, data science, and computing power \hsedit{have enabled} several milestone efforts of unearthing the governing equations for nonlinear dynamical systems. Notably, a breakthrough model named SINDy \hsedit{(Sparse Identification of Nonlinear Dynamics)} \citep{brunton2016discovering} \hsedit{has shed} light on \hsedit{tackling this achallenge}. SINDy was invented to determine the sparse solution among a pre-defined basis function library recursively through a sequential threshold ridge regression (STRidge) algorithm. SINDy quickly became one of the state-of-art methods and kindled significant enthusiasm in this field of study \citep{rudy2017data,long2018pde,champion2019data,chen2021physics,sun2021physics,rao2022discovering}. However, the success of this sparsity-promoting approach relies on a properly defined candidate function library that requires \hsedit{good} prior knowledge of the system. It is also restricted by the fact that a linear combination of candidate functions might be insufficient to recover complicated mathematical expressions. Moreover, when the library size is massive, it empirically fails to hold the sparsity constraint.

At the same time, attempts have been made to tackle the nonlinear dynamics discovery problems by introducing neural networks with activation functions replaced by commonly seen mathematical operators \citep{martius2016extrapolation,sahoo2018learning,kim2019integration,long2019pde}. The intricate formulas are obtained via \hsedit{symbolic expansion of the well-trained network}. This interpretation of \hsedit{physical} laws results in larger candidate pools compared \hsedit{with} the library-based representation of physics employed by SINDy. Nevertheless, since the sparsity of discovered expressions is primarily achieved by empirical pruning of the \hsedit{network} weights, this framework exhibits sensitivity to user-defined thresholds and may fall short to produce parsimonious equations for noisy and scarce data.

Alternatively, another \hsedit{inspiring} work \citep{bongard2007automated,schmidt2009distilling} re-envisioned the data-driven nonlinear dynamics discovery tasks by casting them into symbolic regression problems which have been profoundly resolved by the genetic programming (GP) approach \citep{koza1992genetic,billard2003statistics}. Under this framework, a symbolic regressor is established to identify the governing equations that best describe the \hsedit{underlying physics} through \hsedit{free combination of mathematical operators and symbols, leading to} great flexibility in model selection. One essential weakness of this early methodology is that, driven exclusively by the goal of empirically seeking the best-fitting expression (e.g. minimizing the mean-square error) in a genetic expansion process, the GP-based model usually \hsedit{over-fits} the target system with numerous false-positive terms under data noise, even sometimes at a subtle level, causing huge instability and uncertainty. However, this ingenious idea \hsedit{has inspired} a series of subsequent endeavors \citep{cornforth2012symbolic,gaucel2014learning,ly2012learning,quade2016prediction,vaddireddy2020feature}. In a more recent work, \hsedit{Deep Symbolic Regression (DSR)} \citep{petersen2020deep,mundhenk2021symbolic}, a reinforcement learning-based model was \hsedit{established} and generally outperformed the GP based models including the commercial Eureqa software \citep{langdon2010genetic}. Additionally, the AI-Feynman methods \citep{udrescu2020ai,udrescu2020ai2,udrescu2021symbolic} ameliorated \hsedit{symbolic regression} for distilling physics \hsedit{laws} from data by combining neural network fitting with a suite of physics-inspired techniques. This approach is also highlighted by a recursive decomposition of a complicated mathematical expression into different parts on a tree-based graph, which disentangles the original problem and speeds up the discovery. It outperformed Eureqa in the uncovering Feynman physics equations \citep{feynman1965feynman}. \hsedit{However, this approach is built upon ad-hoc steps and, to some extent, lacks flexible automation in equation discovery.}

The popularity of adopting the tree-based symbolic reasoning of mathematical formulas \citep{lample2019deep} has been rising recently to \hsedit{discover} unknown mathematical expressions with a reinforcement learning agent \citep{kubalik2019symbolic,petersen2020deep,mundhenk2021symbolic}. However, some former work attempting to apply the Monte Carlo tree search (MCTS) algorithm as an alternative to GP for symbolic regression \citep{cazenave2013monte,white2015programming,islam2018expansion,lu2021incorporating} failed to leverage the full flexibility of this algorithm, resulting in the similar shortage \hsedit{that} GP-based symbolic regressors \hsedit{possess} as discussed earlier. Despite these outcomes, we are conscious of the strengths of the MCTS algorithm in equation discovery: it enables the flexible representation of search space with customized computational grammars to guide the search tree expansion. A sound mathematical underpinning for the trade-off between exploration and exploitation is remarkably advantageous as well. These features make it possible to \hsedit{inform} the MCTS agent by \hsedit{our} prior physics knowledge in nonlinear dynamics discovery rather than randomly searching in large spaces. 

\vspace{-6pt}
\paragraph{Contribution.} We propose a promising model named Symbolic Physics Learner (SPL) machine, empowered by MCTS, for discovery of nonlinear dynamics. This architecture relies on a grammar composed of \hsedit{(i)} computational rules and symbols to guide the search tree spanning and \hsedit{(ii)} a composite objective rewarding function to simultaneously evaluate the generated \hsedit{equations} with observational data and enforce \hsedit{the} sparsity of the expression. Moreover, we design multiple adjustments to the conventional MCTS by: \underline{\textbf{(1)}} replacing the expected reward in UCT score with maximum reward to better fit the equation discovery objective, \underline{\textbf{(2)}} employing an adaptive scaling in policy evaluation which would eliminate the uncertainty of the reward value range owing to the unknown error of the system state derivatives, and \underline{\textbf{(3)}} transplanting modules with high returns to the subsequent search as a single leaf node. With these adjustments, the SPL machine is capable of efficiently uncovering the best path to formulate the complex governing equations of the target dynamical system.


\section{Background}

In this section, we expand and explain the background concepts brought up in the introduction to the SPL architecture, including the expression tree (parse tree) and the MCTS algorithm.

\textbf{Expression tree.} Any mathematical expression can be represented by a combinatorial set of symbols and mathematical operations, and further expressed by a parse tree structure \citep{hopcroft2006automata,kusner2017grammar} empowered by a context-free grammar (CFG). A CFG is a formal grammar characterized by a tuple comprised of 4 elements, \hsedit{namely,} $\mathcal{G}=(V, \Sigma, R, S)$, where $V$ denotes a finite set of non-terminal nodes, $\Sigma$ a finite set of terminal nodes, $R$ a finite set of production rules, each interpreted as a mapping from a single non-terminal symbol in $V$ to one or multiple terminal/non-terminal \hsedit{node(s)} in $(V\cup \Sigma )^{*}$ where $*$ represents the Kleene star operation, and $S$ a single non-terminal node standing for a start symbol. In our work, equations are symbolized into parse trees: we define the start node as equation symbol $f$, terminal symbols (leaf nodes) corresponding to the independent variables \hsedit{formulating} the equation (e.g, $x$, $y$), and a \textit{placeholder symbol} $C$ for identifiable constant coefficients that stick to specific production rules. The non-terminal nodes between root and leaf nodes are represented by some symbols distinct from the start and terminal nodes (i.e., $M$). The production rules denote the commonly seen mathematical operators: unary rules (one non-terminal node mapping to one node) for operators like $\cos(\cdot)$, $\exp(\cdot)$, $\log(|\cdot|)$, and binary rules (one non-terminal node mapping to two nodes) for operators such as $+$, $-$, $\times$, $\div$. A parse tree is then generated via a pre-order traversal of production rules rooted at $f$ and terminates when all leaf nodes are entirely filled with terminal symbols. Each mathematical expression can be \hsedit{represented by} such a traversal set of production rules.


\textbf{Monte Carlo tree search.} Monte Carlo tree search (MCTS) \citep{coulom2006efficient} is an algorithm for searching optimal decisions in large combinatorial spaces represented by search trees. This technique complies with the best-first search principle based on the evaluations of stochastic simulations. It has already been widely employed in and proved the spectacular success by various gaming artificial intelligence systems, including the famous AlphaGo and AlphaZero \citep{silver2017mastering} for computer Go game. A basic MCTS algorithm \hsedit{is composed of} an iterative process \hsedit{with} four steps:
\begin{enumerate}
    \item \textbf{Selection.} The MCTS agent, starting from the root node, moves through the visited nodes of the search \hsedit{tree} and selects the next node according to a given selection policy until it reaches an expandable node or a leaf node. 
    \item \textbf{Expansion.} At \hsedit{an}  expandable node, the MCTS agent expands the search tree by selecting one of its unvisited children.  
    \item \textbf{Simulation.} After expansion, if the current node is non-terminal, the agent performs one or multiple independent simulations starting from the current node until reaching the terminal state. In this process, actions are randomly selected. 
    \item \textbf{Backpropagation.} Statistics of nodes along the path from the current node to the root are updated with respect to search results (scores evaluated from the terminate states reached).
\end{enumerate}

To maintain a proper balance between the less-tested paths and the best policy identified so far, the
MCTS agent \hsedit{sticks} to a trade-off between exploration and exploitation by taking action that maximizes the Upper Confidence Bounds applied for Trees (UCT), formulated as \citep{kocsis2006bandit}:
\begin{equation}\label{eq:UCT}
    UCT(s,a) = Q(s,a) + c \sqrt{\ln[N(s)]/N(s,a)}
\end{equation}
where $Q(s,a)$ is the average result/reward of playing action $a$ in state $s$ in the simulations performed in the history, encouraging the exploitation of current best child node; $N(s)$ is number of times state $s$ visited, $N(s, a)$ the number of times action $a$ has been selected at state $s$, and $\sqrt{\ln[N(s)]/N(s,a)}$ consequently encourages exploration of less-visited child nodes. Constant $c$ controls the balance between exploration and exploitation, \hsedit{empirically defined upon the specific problem. Theoretical analysis of UCT-based MCTS (e.g., convergence, guarantees) is referred to \citet{shah2019non}.}

\begin{figure}[t!]
	\centering
	\includegraphics[width=0.99\linewidth]{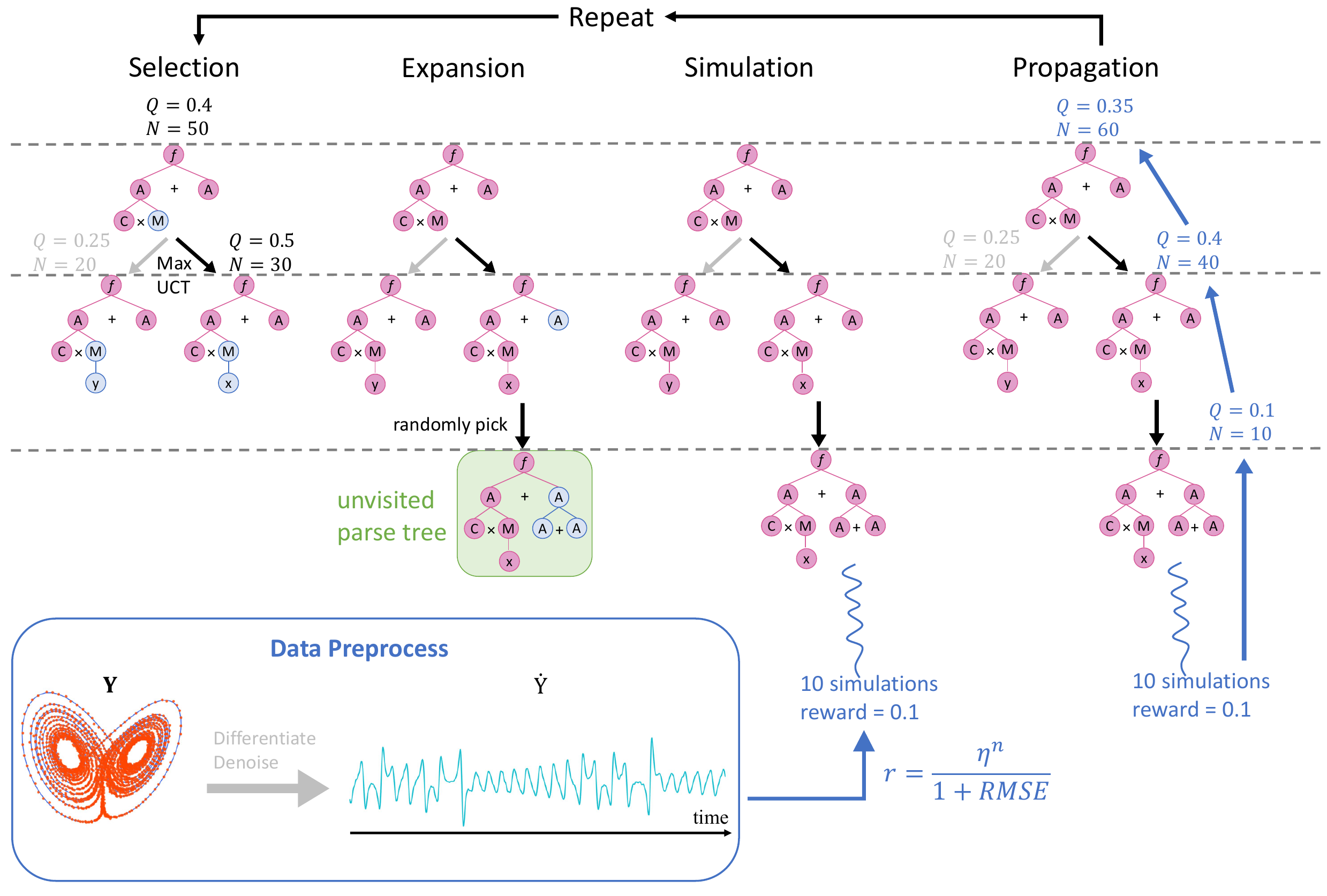}
	\caption{Schematic architecture of the SPL machine for nonlinear dynamics discovery. The graph explains the 4 MCTS phases of one learning episode with an illustrative example. } 
	\label{Figure:SPL}
\end{figure}

\section{Methods}
\hsedit{Existing studies show} that the MCTS agent continuously gains knowledge of \hsedit{specified tasks} via the expansion of the search tree and, based on the backpropagation of evaluation results (i.e., rewards and number of visits), render a proper selection policy on visited states to guide the upcoming searching \hsedit{\citep{silver2017mastering}}. In the proposed SPL machine, \hsedit{such a} process is integrated with the symbolic reasoning of mathematical expressions to reproduce and evaluate valid mathematical expressions of the \hsedit{physical} laws in nonlinear dynamics step-by-step, and then obtain a favorable selection policy pointing to the best solution. This algorithm is depicted in Figure \ref{Figure:SPL} with an illustrative example and its overall training scheme is shown in Algorithm \ref{alg:SPL}. \hsedit{Discussion of the hyperparameter setting for this algorithm is given in Appendix Section \ref{appendix:hyperparameter}.}

\textbf{Rewarding.} To evaluate the mathematical expression $\Tilde{f}$ projected from a parse tree, we define a numerical reward $r \in \mathcal{R} \subset \mathbb{R}$ based on this expression and input data $\mathcal{D}=\{\mathbf{Y}; \dot{Y}_i\}$, serving as the search result of the current expansion or simulation. It is formulated as
\begin{equation}\label{eq:reward}
  r=\frac{\eta^n}{1+\sqrt{\frac{1}{N}\big\|\dot{Y}_i-\Tilde{f}(\mathbf{Y})\big\|_2^2}}
\end{equation}
where $\mathbf{Y}=\{\mathbf{y}_1, \mathbf{y}_2, ..., \mathbf{y}_m\} \in \mathbb{R}^{m \times N}$ is the $m$ dimensional state variables \hsedit{of a dynamical system}, $\dot{Y}_i \in \mathbb{R}^{1 \times N}$ the numerically estimated state derivative for $i$th dimension, and $N$ the number of measurement data points. $\eta$ denotes a discount factor, assigned slightly smaller than 1; $n$ is empirically defined as the total number of production rules in the parse tree. This numerator arrangement is designated to penalize non-parsimonious solutions. This rewarding formulation outputs a reasonable assessment to the distilled equations and encourages parsimonious solution by discounting the reward of a non-parsimonious one. The rooted mean square error (RMSE) in denominator evaluates the goodness-of-fit of the discovered equation w.r.t. the measurement data.

\begin{algorithm}[t!] 
\small
\SetAlgoLined
    \textbf{Input:} Grammar $G=(V, \Sigma, R, S)$, measurement data $\mathcal{D}=\{\mathbf{Y}; \dot{Y}_i\}$\;
    \textbf{Parameters:} \hsedit{discount/regularization} factor $\eta$, exploration rate $c$, $t_{max}$;~~~~~{\color{gray}\hsedit{\# $\eta$ controls equation parsimony}}\;
    \textbf{Output:} Optimal governing equation $\tilde{f}^\star$\;
    \For{\textbf{each episode}}{
        \textbf{Selection:} Initialize $s_0=\emptyset, t=0, NT=[S]$\;
        \While {$s_t$ expandable and $t < t_{max}$}{
            Choose $a_{t+1}= \argmax_{\mathcal{A}} UCT(s_t, a)$\;
            Take action $a_{t+1}$, observe $s', NT$\;
            $s_{t+1} \leftarrow s'$ note as visited, $t \leftarrow t+1$\;
        }
        \textbf{Expansion:} Randomly take an unvisited path with action $a$, observe $s', NT$\;
        $s_{t+1} \leftarrow s'$ note as visited, $t \leftarrow t+1$\;
        \If{$NT = \emptyset$}{
            Project $\Tilde{f}$, \textbf{Backpropagate} $r_{t+1}$ and visited count and finish the episode\; 
        }
        \textbf{Simulation:} Fix the starting point $s_t, NT$\;
        \For {\textbf{each simulation}}{
            \While {$s_t$ non-terminal and $t < t_{max}$}{
                Randomly take an action $a$, observe $s', NT$\;
                $s_{t+1} \leftarrow s', t \leftarrow t+1$\;
            }
            \If{$NT = \emptyset$}{
                Project $\Tilde{f}$ and calculate $r_{t+1}$\;
            }
        }
        \textbf{Backpropagate} simulation results\; 
    }\caption{\hsedit{Training SPL for discovering the $i^{th}$ governing equation ($i=1,2,...,m$)}}\label{alg:SPL}
\end{algorithm}

\textbf{Training scheme.} A grammar $\mathcal{G}=(V, \Sigma, R, S)$ is defined with appropriate nodes and production rules to cover all possible forms of equations. To keep track of non-terminal nodes of the parsing tree, we apply a last-in-first-out (LIFO) strategy and denote the non-terminal node placed last on the stack $NT$ as the current node. We define the action space $\mathcal{A}=R$ and the state space $\mathcal{S}$ as all possible traversals of complete/incomplete parse trees (i.e., production rules selected) in ordered sequences. At the current state $s_t=[a_1, a_2, ...a_t]$ where $t \in \mathbb{N}$ is the discrete traversal step-index of the upcoming production rule, the MCTS agent masks out the invalid production rules for current non-terminal node and on that basis selects a valid rule as action $a_{t+1}$  (i.e, the left-hand side of a valid production rule is the current non-terminal symbol). Consequently, the parse tree gains a new terminal/non-terminal branch in accordance with $a_{t+1}$, meanwhile the agent finds itself in a new state $s_{t+1}=[a_1, a_2, ...a_t, a_{t+1}]$. The agent subsequently pops off the current non-terminal symbol from $NT$ and pushes the non-terminal nodes, if there are any, on the right-hand side of the selected rule onto the stack. Once the agent attains an unvisited node, a certain amount of simulations are performed, where the agent starts to randomly select the next node until the parse tree is completed. The reward is calculated or the maximal size is exceeded, resulting in a zero reward. The best result from the attempts counts as the reward of the current simulation phase and backpropagates from the current unvisited node all the way to the root node.

\textbf{Greedy search.} Different from the MCTS-based gaming AIs where the agents are inclined to pick the action with a high expected reward (average returns), the SPL machine seeks the unique optimal solution. In the proposed training framework, we apply a greedy search heuristic to encourage the agent to explore the branch which yields the best solution in the past: $Q(s,a)$ is defined as the maximum reward of the state-action pair, and its value is backpropagated from the highest reward in the simulations upon the selection of the pair. Meanwhile, to overcome the local minima problems due to this greedy approach in policy search, we enforce a certain level of randomness by empirically adopting the $\epsilon$-greedy algorithm, a commonly seen approach in reinforcement learning models.

\textbf{Adaptive-scaled rewarding.} Owing to the unknown level of error from the numerically estimated state derivatives, the range of the RMSE in the SPL reward function is \hsedit{unpredictable}. This uncertainty affects the scale of rewarding values thus the balance between exploration and exploitation is presented in Eq. \eref{eq:UCT}. Besides adding ``1'' to the denominator of Eq. \eref{eq:reward} to avoid dramatically large numerical rewards, we also apply an adaptive scale of the reward, \hsedit{given by}
\begin{equation}\label{eq:c}
    Q(s,a) = \frac{r^*(s,a)}{\max_{s' \in \mathcal{S}, a' \in \mathcal{A}} Q(s',a')}
\end{equation}
where $r^*$ denotes the maximum reward of the state-action pair. It is scaled by the current maximum reward among all states $\mathcal{S}$ and actions $\mathcal{A}$ to reach an equilibrium that the $Q$-values are stretched to the scale $[0, 1]$ at any time. This self-adaptive fashion yields a well-scaled calculation of UCT under different value ranges of rewards throughout the training.

\begin{wrapfigure}[12]{r}{0.7\textwidth}
\vspace{-8pt}
\begin{center}
\includegraphics[width=0.99\linewidth]{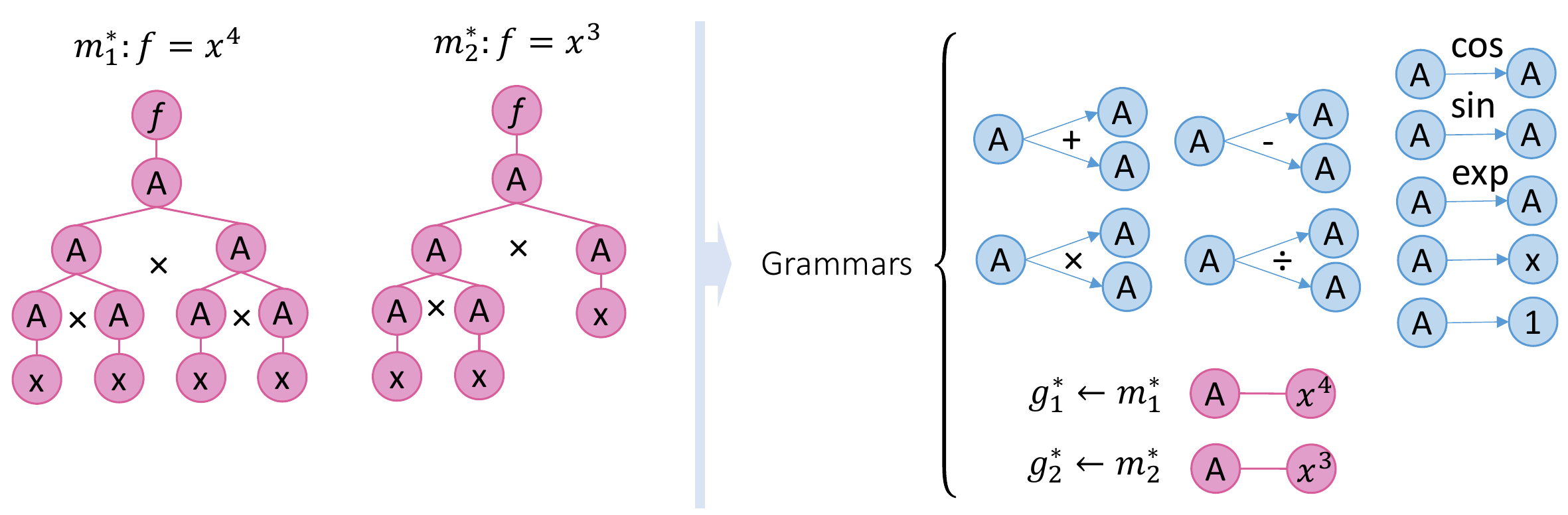}
\vspace{-9pt}
\caption{A module transplantation process: A complete parse serves as a single production rule and is appended to the grammar pool.}
\label{Figure:module}
\end{center}
\vspace{-12pt}
\end{wrapfigure}

\textbf{Module transplantation.} A function can be decomposed into smaller modules where each is simpler than the original one \citep{udrescu2020ai2}. This modularity feature, as shown in Figure \ref{Figure:module}, helps us develop a divide-and-conquer heuristic for distilling some complicated functions: after every certain amount of MCTS iterations, the discovered parse trees with high rewards are picked out and thereupon 
\normalsize
reckoned as individual production rules and appended to the set of production rules $R$; accordingly, these trees are ``transplanted'' to the future ones as their modules (i.e, the leaves). To avoid early overfitting problem, we incrementally enlarge the sizes of such modules from a baseline length to the maximum allowed size of the parse tree throughout the iterations. The augmentation part of $R$ is refreshed whenever new production rules are created, keeping only the ones engendering high rewards. This approach accelerates the policy search by capturing and locking some modules that likely contribute to, or appear as part of the optimal solution, especially in the cases of the mathematical expression containing ``deep'' operations (e.g., high-order polynomials) whose structures are difficult for the MCTS agent to repeatedly obtain during the selection and expansion.

\section{Symbolic Regression: Finding Mathematical Formulas}

\subsection{Data Noise \& Scarcity}\label{data_scarcity}

\begin{wrapfigure}[13]{r}{0.6\textwidth}
\vspace{-40pt}
\begin{center}
\includegraphics[width=0.95\linewidth]{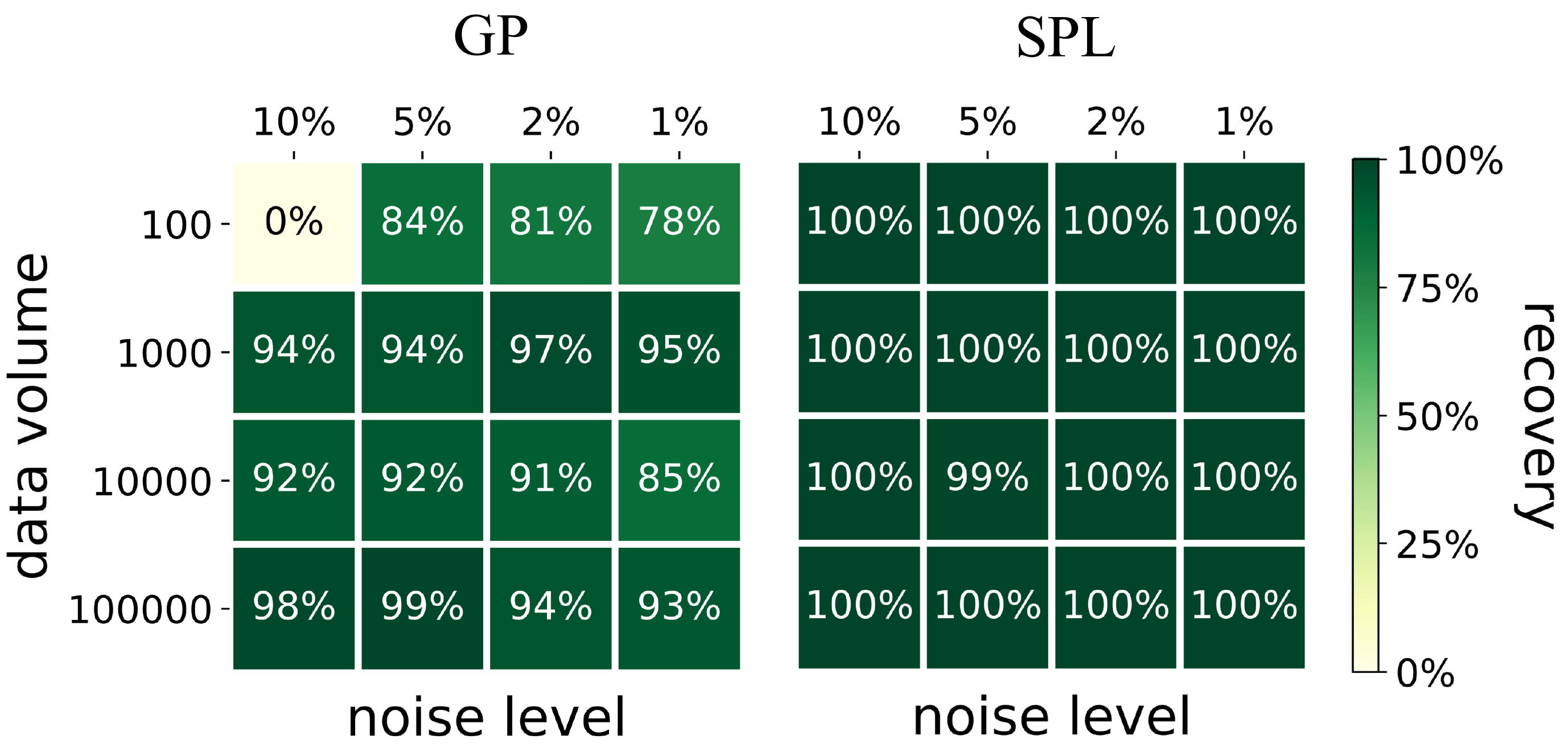}
\vspace{-6pt}
\caption{The effect of data noise/scarcity on recovery rate. The heatmaps demonstrate the recovery rate of GP-based symbolic regressor and the SPL machine under different data conditions, summarized over 100 independent trials.}
\label{Figure:SN}
\end{center}
\vspace{-12pt}
\end{wrapfigure}
\normalsize

Data scarcity and noise are commonly seen in measurement data and become one of the bottleneck issues for discovering the governing equations of nonlinear dynamics. Tackling the challenges in high-level data scarcity and noise situations is traditionally regarded as an essential robustness indicator for a nonlinear dynamics discovery model. \hsedit{To this end}, we present an examination of the proposed SPL machine by an equation discovery task in the presence of multiple levels of data noise and volume, comparing with a GP-based symbolic regressor (implemented with \textit{gplearn} python package)\footnote{All simulations are performed on a \hsedit{standard} workstation with a NVIDIA GeForce RTX 2080Ti GPU.}. The target equation is $f(x)=0.3x^3+0.5x^2+2x$, and the independent variable $X$ is uniformly sampled in the given range $[-10, 10]$. Gaussian white noise is added to the dependent variable $Y$ with the noise level defined as the root-mean-square ratio between the noise and the exact values. For discovery, the two models are fed with equivalent search space: $\{+, -, \times, \div, cost, x\}$ as candidate mathematical operations and symbols. The hyperparameters of \hsedit{the} SPL machine are set as $\eta=0.99$, $t_{max}=50$, and 10,000 episodes of training is regarded as one trail. For the GP-based symbolic regressor, \hsedit{the} population of programs is set \hsedit{as 2,000}, the number of generations as 20. \hsedit{The} range of constant \hsedit{coefficient} values is $[-10, 10]$. For 16 different data noise and scarcity levels, each model \hsedit{was performed} 100 independent trails. The recovery rates are displayed as a $4\times 4$ mesh grid w.r.t. different noise/scarcity levels in Figure \ref{Figure:SN}. It is observed that the SPL machine outperforms the GP-based symbolic regressor in all the cases. A T-test also proves that \hsedit{the} recovery rate of the SPL machine is significantly higher than that of GP (\hsedit{e.g.,} $p$-value = $1.06\times 10^{-7}$).



\subsection{Nguyen's Symbolic Regression Benchmark}\label{ss:SR1}

\begin{table}[t!]
    \centering
    \caption{Recovery rate of three algorithms in Nguyen's benchmark symbolic regression problems. The SPL machine outperforms the other two models in average recovery rate. }
    \vspace{6pt}
    \small
    \begin{tabular}{ccccc}
    \toprule
    \textbf{Benchmark} & \textbf{Expression}  &  \textbf{SPL} &  \textbf{NGGP} & \textbf{GP} \\
    \midrule
    Nguyen-1 & $x^3+x^2+x$                      & 100\% & 100\% & 99\% \\ 
    Nguyen-2 & $x^4+x^3+x^2+x$                  & 100\% & 100\% & 90\% \\
    Nguyen-3 & $x^5+x^4+x^3+x^2+x$ 	            & 100\% & 100\% & 34\% \\
    Nguyen-4 & $x^6+x^5+x^4+x^3+x^2+x$          & 99\% & 100\% & 54\%\\
    Nguyen-5 & $\sin(x^2)\cos(x)-1$             & 95\% & 80\% & 12\%\\
    Nguyen-6 & $\sin(x^2)+\sin(x+x^2)$	        & 100\% & 100\% & 11\%\\
    Nguyen-7 & $\ln(x+1)+\ln(x^2+1)$            & 100\% & 100\% & 17\%\\
    Nguyen-8 & $\sqrt{x}$                       & 100\% & 100\% & 100\%\\
    Nguyen-9 & $\sin(x)+\sin(y^2)$ 	            & 100\% & 100\% & 76\%\\
    Nguyen-10 & $2\sin(x)\cos(y)$               & 100\% & 100\% & 86\%\\
    Nguyen-11 & $x^y$                           & 100\% & 100\% & 13\%\\
    Nguyen-12 & \hsedit{$x^4-x^3+\frac{1}{2}y^2-y$}	    & 28\% & 4\% & 0\%\\
    \midrule
    Nguyen-1$^c$ & $3.39x^3 + 2.12x^2 + 1.78x$  & 100\% & 100\% & 0\% \\ 
    Nguyen-2$^c$ & $0.48x^4 + 3.39x^3 + 2.12x^2 + 1.78x$ & 94\% & 100\% & 0\% \\ 
    Nguyen-5$^c$ & $\sin(x^2)\cos(x)-0.75$      & 95\% & 98\% & 1\% \\
    Nguyen-8$^c$ & $\sqrt{1.23x}$               & 100\% & 100\% & 56\% \\
    Nguyen-9$^c$ & $\sin(1.5x)+\sin(0.5y^2)$ 	& 96\% & 90\% & 0\% \\
    \midrule
    Average &                                   & \textbf{94.5\%} & 92.4\% & 38.2\%\\
    \bottomrule
    \end{tabular}
    \label{Table:SR}
    \vspace{-6pt}
    \normalsize
\end{table}

Nguyen's symbolic regression benchmark task \citep{uy2011semantically} is widely used \hsedit{to test} the model's robustness in symbolic regression problems. Given a set of allowed operators, a target equation, and data generated by \hsedit{the specified} equation \hsedit{(see Table \ref{Table:SR} for example)}, the tested model is supposed to distill \hsedit{the} mathematical expression that is identical to the target equation, or equivalent to it (e.g., Nguyen-7 equation can be recovered as $\log(x^3 + x^2 + x + 1)$, Nguyen-10 equation can be recovered as $\sin(x+y)$, and Nguyen-11 equation can be recovered as $\exp(y\log(x))$). Some variants of Nguyen's benchmark equations are also considered in this experiment. Their discoveries require numerical estimation of \hsedit{the constant coefficient} values. Each equation generates two datasets: one for training and another for testing. The discovered equation that perfectly fits the testing data is regarded as a successful discovery \hsedit{(i.e., the discovered equation should be identical or equivalent to the target one)}. The recovery rate is calculated based on 100 independent tests for each task. In these benchmark tasks, three algorithms are tested: GP-based symbolic regressor, the neural-guided GP (NGGP) \citep{mundhenk2021symbolic}, and the SPL machine. Note that NGGP is an improved approach over DSR \citep{petersen2020deep}. They are given the same set of candidate operations: $\{+,-,\times,\div,\exp(\cdot),\cos(\cdot),\sin(\cdot)\}$ for all benchmarks and $\{\sqrt{\cdot},\ln(\cdot)\}$ are added to the 7, 8, 11 benchmarks. The hyperparameters of the GP-based symbolic regressor are the same as those mentioned in Section \ref{data_scarcity}; configurations of the NGGP models are obtained from its source code\footnote{NGGP source code: https://github.com/brendenpetersen/deep-symbolic-optimization/tree/master/dso/dso}; detailed setting of the benchmark tasks and the SPL model is described in {\color{black} Appendix Section \ref{appendix:nguyen}}. The success rates are shown in Table \ref{Table:SR}. It is observed that the SPL machine and the NGGP model both produce reliable results in Nguyen's benchmark problems and the SPL machine slightly \hsedit{outperforms} NGGP. This experiment betokens the capacity of the SPL machine in \hsedit{discovery of equations} with divergent forms. 

\textbf{Ablation Study:} We consider \hsedit{four} ablation studies by removing: (a) the adaptive scaling in reward calculation, (b) the discount factor $\eta^n$ that drives equation parsimony in Eq. \eref{eq:reward}, (c) module transplantation in tree generation, \hsedit{and (d) all of the above}. The \hsedit{four} models were tested on the \hsedit{first 12} Nguyen equations \hsedit{(see Appendix Section \ref{appendix:ablation})}. Results show the average recovery rates for these models are all \hsedit{smaller than that produced by SPL (see Appendix Table \ref{Table:Appendix_ablation}), where the module transplantation brings the largest effect}. Hence, these modules are critical to guarantee the proposed model efficacy.

\vspace{12pt}
\section{\hsedit{Physical Law} Discovery: Free Falling Balls with Air Resistance}

\begin{wraptable}[7]{r}{0.53\textwidth}
\vspace{-22pt}
\caption{Baseline models ($c_i$: unknown constants).}\label{Table:DropModel}
\centering
\vspace{6pt}
\small
\begin{tabular}{ll}
\toprule
\textbf{Physics Model} & \textbf{Derived model expression}\\
\midrule
Model-1  & $H(t)=c_0 + c_1 t + c_2 t^2 + c_3 t^3$\\
Model-2 & $H(t)=c_0 + c_1 t + c_2 e^{c_3 t}$  \\
Model-3 & $H(t)=c_0 + c_1\log(\cosh(c_2 t))$\\
\bottomrule
\end{tabular}
\normalsize
\end{wraptable} 

It is well known that, in 1589–1592, Galileo dropped two objects of unequal mass from the Leaning Tower of Pisa and drew a conclusion that their velocities were not affected by the mass. This has been well recognized globally as the ``textbook'' \hsedit{physical law} for the vertical motion of a free-falling object: the height of the object is formulated as $H(t) = h_0 + v_0 t - \frac{1}{2}gt^2$, where $h_0$ denotes initial height, $v_0$ the initial velocity, and $g$ the gravitational acceleration. However, this ideal situation is rarely reached in our daily life because air resistance serves as a significant damping factor that prevents the above \hsedit{physical law} from occurring in real-life cases.

\begin{wraptable}[16]{r}{0.63\textwidth}
\vspace{-22pt}
\caption{Mean square error (MSE) between ball motion prediction with the measurements in the test set. The SPL machine reaches the best prediction results in most (9 out of 11) cases.}\label{Table:ball mse}
\centering
\vspace{6pt}
\small
\begin{tabular}{ccccc}
\toprule
\textbf{Type} &  \textbf{SPL} &  \textbf{Model-1} & \textbf{Model-2} & \textbf{Model-3} \\
\midrule
baseball             & \textbf{0.3} & 2.798 & 94.589 & 3.507\\
blue basketball      & \textbf{0.457} & 0.513 & 69.209 & 2.227\\
green basketball     & \textbf{0.088} & 0.1 & 85.435 & 1.604\\
volleyball           & \textbf{0.111} & 0.574 & 80.965 & 0.76\\
bowling ball         & \textbf{0.003} & 0.33 & 87.02 & 3.167\\
golf ball            & \textbf{0.009} & 0.214 & 86.093 & 1.684\\
tennis ball          & \textbf{0.091} & 0.246 & 72.278 & 0.161\\
whiffle ball 1       & 1.58 & 1.619 & 65.426 & \textbf{0.21}\\
whiffle ball 2       & \textbf{0.099} & 0.628 & 58.533 & 0.966\\
yellow whiffle ball  & \textbf{0.428} & 17.341 & 44.984 & 2.57\\
orange whiffle ball  & 0.745 & \textbf{0.379} & 36.765 & 3.257\\
\bottomrule
\end{tabular}
\normalsize
\end{wraptable} 

Many efforts have been made to uncover the effect of the air resistance and derive mathematical models to describe the free-falling objects with air resistance \citep{clancy1975aerodynamics,lindemuth1971effect,greenwood1986air}. This section provides data-driven discovery of the \hsedit{physical} laws of relationships between height and time in the cases of free-falling objects with air resistance based on multiple experimental ball-drop datasets \citep{de2020discovery}, which contain the records of 11 different types of balls dropped from a bridge (see {\color{black} Appendix Figure \ref{Figure:balls}}). For discovery, each dataset is split into a \hsedit{training} set (records from the first 2 seconds) and a \hsedit{testing} set (records after 2 seconds). Three mathematically derived physics models are selected from the literature as baseline models\footnote{Model 1: https://faraday.physics.utoronto.ca/IYearLab/Intros/FreeFall/FreeFall.html}$^,$\footnote{Model 2: https://physics.csuchico.edu/kagan/204A/lecturenotes/Section15.pdf}$^,$\footnote{Model 3: https://en.wikipedia.org/wiki/Free\_fall} for this experiment (see Table \ref{Table:DropModel}), and the unknown constant \hsedit{coefficient} values are \hsedit{estimated} by Powell's conjugate direction method \citep{powell1964efficient}. Based on \hsedit{our} prior knowledge of the \hsedit{physical} law that may appear in this case, \hsedit{we} use $\{+,-,\times,\div,\exp(\cdot),\cosh(\cdot),\log(\cdot)\}$ as the candidate grammars \hsedit{for} the SPL discovery, with terminal nodes $\{t, const\}$. The hyperparameters are set as $\eta=0.9999$, $t_{max}=20$, and one single discovery is built upon 6,000 episodes of training. The \hsedit{physical} laws distilled by SPL from training data are applied to the test data and compared with the ground truth. Their prediction errors, in terms of MSE, are presented in Table \ref{Table:ball mse} (the SPL-discovered equations are shown in Appendix Table \ref{Table:ball eq}). The full results can be found in {\color{black} Appendix Section \ref{appendix:free falling}}. It can be concluded that the data-driven discovery of physical laws leads to a better approximation of the free-falling objects with air resistance.

\vspace{-6pt}
\section{Chaotic Dynamics Discovery: The Lorenz System}

The first nonlinear dynamics discovery example is a 3-dimensional Lorenz system \citep{lorenz1963deterministic} whose dynamical behavior ($x, y, z$) is governed by $\dot{x} = \sigma(y-x), \dot{y} = x(\rho-z)-y, \dot{z} = xy-\beta z$ with parameters $\sigma=10$, $\beta=8/3$, and $\rho=28$. \hsedit{The} Lorenz attractor has two lobes and the system, starting from anywhere, makes cycles around one lobe before switching to the other and iterates repeatedly, \hsedit{exhibiting} strong chaos. The synthetic \hsedit{system states ($x, y, z$)} are generated by solving the nonlinear differential equations using \hsedit{the} Matlab \texttt{ode113} \citep{shampine1975computer,shampine1997matlab} function. 5\% Gaussian white noise is added to the clean data to generate noisy measurement. The derivatives of the system states \hsedit{($\dot{x}, \dot{y}, \dot{z}$)} are \hsedit{unmeasured but} estimated by central difference and smoothed by the Savitzky–Golay filter \citep{savitzky1964smoothing} in order to reduce the noise effect.

\begin{wraptable}[13]{r}{0.43\textwidth}
\vspace{-10pt}
\caption{Summary of the discovered governing equations for Lonrez system. Each cell concludes if target physics terms are distilled (if yes, number of false positive terms in uncovered expression).}\label{Table:Lorenz1}
\centering
\vspace{6pt}
\small
{
\begin{tabular}{llll}
\toprule
\textbf{Model} & \textbf{$\dot{x}$} &\textbf{$\dot{y}$} & \textbf{$\dot{z}$}\\
\midrule
Eureqa  & Yes (1) & Yes (3) & Yes (1)\\
\midrule
pySINDy & Yes (1) & No (N/A) & Yes (2)\\
\midrule
NGGP & Yes (10) & Yes (8) & Yes (16)\\
\midrule
SPL     & \textbf{Yes (0)} & \textbf{Yes (0)} & \textbf{Yes (0)}\\
\bottomrule
\end{tabular}}
\normalsize
\end{wraptable}

In this experiment, the proposed SPL machine \hsedit{is compared} with three benchmark methods: Eureqa, pySINDy and NGGP. For Eureqa, NGGP, and the SPL machine, $\{+, -, \times, \div \}$ are used as candidate operations; the upper bound of complexity is set to be 50; for pySINDy, the candidate function library includes all polynomial basis of \hsedit{($x, y, z$)} from degree 1 to degree 4. {\color{black} Appendix Table \ref{Table:Lorenz}} presents the distilled governing \hsedit{equations} by each approach and Table \ref{Table:Lorenz1} summarizes these results: the SPL machine uncovers the explicit form of equations accurately in the context of active terms, whereas Eureqa, pySINDy and NGGP yield several false-positive terms in the governing equations. In particular, although Eureqa and NGGP are capable of uncovering the correct terms, their performance is very sensitive to the measurement noise as indicated by the redundant terms (despite with small coefficients) shown in Appendix Table \ref{Table:Lorenz}. Overall, the baseline methods fail to handle the large noise effect, essentially limiting their applicability in nonlinear dynamics discovery. It is evident that the SPL machine is capable of distilling the concise symbolic combination of operators and variables to correctly formulate parsimonious mathematical expressions that govern the Lorenz system, outperforming the baseline methods of Eureqa, pySINDy and NGGP.

\section{Experimental Dynamics Discovery: Double Pendulum}

This section shows SPL-based discovery of a chaotic double pendulum system with experimental data \citep{asseman2018learning} as shown in Appendix Figure \ref{Figure:DP}. The governing equations \hsedit{are} given by: 
\begin{equation} \label{ch4:eq:DQ2} 
\begin{split}
    \dot{\omega}_1&=c_{1}\dot{\omega}_2\cos(\Delta\theta)+c_{2}\omega_2^2\sin(\Delta\theta) + c_{3}\sin(\theta_1)+ \mathcal{R}_1(\theta_1, \theta_2, \dot{\theta}_1, \dot{\theta}_2), \\
    \dot{\omega}_2&=c_{1}\dot{\omega}_1\cos(\Delta\theta)+c_{2}\omega_1^2\sin(\Delta\theta)+c_{3}\sin(\theta_2)+\mathcal{R}_2(\theta_1, \theta_2, \dot{\theta}_1, \dot{\theta}_2)
\end{split}
\end{equation}
where ${\theta}_1$, ${\theta}_2$ denote the angular displacements; $\omega_1=\dot{\theta}_1$, $\omega_2=\dot{\theta}_2$ the velocities; $\dot{\omega}_1$, $\dot{\omega}_2$ the accelerations; $\mathcal{R}_1$ and $\mathcal{R}_2$ denote the unknown damping forces. Note that $\Delta\theta=\theta_1 - \theta_2$. 


The data source contains multiple camera-sensed datasets. Here, 5,000 denoised random sub-samples from 5 datasets are used for training, 2,000 random sub-samples from another 2 datasets for validation, and 1 dataset for testing. The derivatives of the system states are numerically estimated by the same approach discussed in the Lorenz case. Some \hsedit{prior physics knowledge} is employed to guide the discovery: (\textbf{1}) \hsedit{the terms} $\dot{\omega}_2\cos(\Delta\theta)$ for $\dot{\omega}_1$ and $\dot{\omega}_1\cos(\Delta\theta)$ \hsedit{for $\dot{\omega}_2$} are \hsedit{assumed to be part of the governing equations based on the Lagrange derivation}; (\textbf{2}) the angles ($\theta_1$, $\theta_2$, $\Delta\theta$) are under the trigonometric functions $\cos(\cdot)$ and $\sin(\cdot)$; (\textbf{3}) directions of velocities/relative velocities may appear in damping. Production rules fulfilling the above prior knowledge are exhibited in {\color{black} Appendix Section \ref{appen:DoublePendulum}}. The hyperparameters are set as $\eta=1$, $t_{max}=20$, and 40,000 episodes of training \hsedit{are} regarded as one trail. 5 independent trials are performed and the equations with the highest validation scores are selected as the final results. The uncovered equations are given as follows: 
\begin{equation} \label{ch4:eq:DQ2} 
\begin{split}
    \dot{\omega}_1&=-0.0991\dot{\omega}_2\cos(\Delta\theta)-0.103\omega_2^2\sin(\Delta\theta)-69.274\sin(\theta_1)+ \underline{0.515\cos(\theta_1)}, \\
    \dot{\omega}_2&=-1.368\dot{\omega}_1\cos(\Delta\theta)+1.363\omega_1^2\sin(\Delta\theta)-92.913\sin(\theta_2)+ \underline{0.032\omega_1},
\end{split}
\end{equation}
where the explicit expression of physics in an ideal double pendulum system, as displayed in Eq. \eref{ch4:eq:DQ2}, are successfully distilled and damping terms are estimated. This set of equation is validated through interpolation on the testing set and compared with the smoothed derivatives, as shown in Appendix Figure \ref{Figure:s:dp_pred}. The solution appears felicitous as the governing equations of the testing responses.

\section{Conclusion and Discussion}\label{s:conclusion}

This paper introduces a Symbolic Physics Learner (SPL) machine to tackle the challenge of distilling the mathematical structure of equations \hsedit{for physical systems (e.g., nonlinear dynamics)} with \hsedit{scarce/noisy} data. This framework is built upon the expression tree interpretation of mathematical operations and variables and an MCTS agent that searches for the optimistic policy to reconstruct the target mathematical \hsedit{formula}. With some remarkable adjustments to the MCTS algorithms, the SPL model straightforwardly accepts \hsedit{our} prior or domain knowledge, or any \hsedit{sort} of constraints of the tasks in the grammar design while leveraging great flexibility in expression formulation. The robustness of the proposed SPL \hsedit{machine} for complex target expression discovery \hsedit{within a} large search space is indicated in \hsedit{the} Nguyen's symbolic regression benchmark problems, where the SPL machine outperforms state-of-the-art symbolic regression methods. Moreover, encouraging results are obtained in the \hsedit{tasks of discovering physical laws} and nonlinear dynamics, based on synthetic or experimental datasets. While the proposed SPL \hsedit{machine} shows huge potential in both symbolic regression and \hsedit{physical law} discovery tasks, there are still some imperfections that can be improved: (i) the computational cost is high for constant \hsedit{coefficient} value estimation \hsedit{due to repeated calls for an optimization process}, (ii) graph modularity is underexamined, and (iii) robustness against extreme data noise and scarcity is not optimal. These limitations are further explained in {\color{black} Appendix Section \ref{appendix:discussion}}.


\subsubsection*{Acknowledgments}
The work is supported by the National Natural Science Foundation of China (No. 92270118) and the Beijing Outstanding Young Scientist Program (No. BJJWZYJH012019100020098).

\bibliography{iclr2023_conference}

\begin{thebibliography}{52}
\providecommand{\natexlab}[1]{#1}
\providecommand{\url}[1]{\texttt{#1}}
\expandafter\ifx\csname urlstyle\endcsname\relax
  \providecommand{\doi}[1]{doi: #1}\else
  \providecommand{\doi}{doi: \begingroup \urlstyle{rm}\Url}\fi

\bibitem[Asseman et~al.(2018)Asseman, Kornuta, and Ozcan]{asseman2018learning}
Alexis Asseman, Tomasz Kornuta, and Ahmet Ozcan.
\newblock Learning beyond simulated physics.
\newblock In \emph{Modeling and Decision-making in the Spatiotemporal Domain
  Workshop--NIPS}, 2018.

\bibitem[Billard \& Diday(2003)Billard and Diday]{billard2003statistics}
L~Billard and E~Diday.
\newblock From the statistics of data to the statistics of knowledge: symbolic
  data analysis.
\newblock \emph{Journal of the American Statistical Association}, 98\penalty0
  (462):\penalty0 470--487, 2003.

\bibitem[Bongard \& Lipson(2007)Bongard and Lipson]{bongard2007automated}
Josh Bongard and Hod Lipson.
\newblock Automated reverse engineering of nonlinear dynamical systems.
\newblock \emph{Proceedings of the National Academy of Sciences}, 104\penalty0
  (24):\penalty0 9943--9948, 2007.

\bibitem[Brunton et~al.(2016)Brunton, Proctor, and
  Kutz]{brunton2016discovering}
Steven~L Brunton, Joshua~L Proctor, and J~Nathan Kutz.
\newblock Discovering governing equations from data by sparse identification of
  nonlinear dynamical systems.
\newblock \emph{Proceedings of the national academy of sciences}, 113\penalty0
  (15):\penalty0 3932--3937, 2016.

\bibitem[Cazenave(2013)]{cazenave2013monte}
Tristan Cazenave.
\newblock Monte-carlo expression discovery.
\newblock \emph{International Journal on Artificial Intelligence Tools},
  22\penalty0 (01):\penalty0 1250035, 2013.

\bibitem[Champion et~al.(2019)Champion, Lusch, Kutz, and
  Brunton]{champion2019data}
Kathleen Champion, Bethany Lusch, J~Nathan Kutz, and Steven~L Brunton.
\newblock Data-driven discovery of coordinates and governing equations.
\newblock \emph{Proceedings of the National Academy of Sciences}, 116\penalty0
  (45):\penalty0 22445--22451, 2019.

\bibitem[Chen et~al.(2021)Chen, Liu, and Sun]{chen2021physics}
Zhao Chen, Yang Liu, and Hao Sun.
\newblock Physics-informed learning of governing equations from scarce data.
\newblock \emph{Nature Communications}, 12:\penalty0 6136, 2021.

\bibitem[Clancy(1975)]{clancy1975aerodynamics}
Laurence~Joseph Clancy.
\newblock \emph{Aerodynamics}.
\newblock John Wiley \& Sons, 1975.

\bibitem[Cornforth \& Lipson(2012)Cornforth and Lipson]{cornforth2012symbolic}
Theodore Cornforth and Hod Lipson.
\newblock Symbolic regression of multiple-time-scale dynamical systems.
\newblock In \emph{Proceedings of the 14th annual conference on Genetic and
  evolutionary computation}, pp.\  735--742, 2012.

\bibitem[Coulom(2006)]{coulom2006efficient}
R{\'e}mi Coulom.
\newblock Efficient selectivity and backup operators in monte-carlo tree
  search.
\newblock In \emph{International conference on computers and games}, pp.\
  72--83. Springer, 2006.

\bibitem[de~Silva et~al.(2020)de~Silva, Higdon, Brunton, and
  Kutz]{de2020discovery}
Brian~M de~Silva, David~M Higdon, Steven~L Brunton, and J~Nathan Kutz.
\newblock Discovery of physics from data: universal laws and discrepancies.
\newblock \emph{Frontiers in artificial intelligence}, 3:\penalty0 25, 2020.

\bibitem[Dzeroski \& Todorovski(1995)Dzeroski and
  Todorovski]{dzeroski1995discovering}
Saso Dzeroski and Ljupco Todorovski.
\newblock Discovering dynamics: from inductive logic programming to machine
  discovery.
\newblock \emph{Journal of Intelligent Information Systems}, 4\penalty0
  (1):\penalty0 89--108, 1995.

\bibitem[D{\v{z}}eroski \& Todorovski(1993)D{\v{z}}eroski and
  Todorovski]{dvzeroski1993discovering}
Sa{\v{s}}o D{\v{z}}eroski and Ljup{\'e}o Todorovski.
\newblock Discovering dynamics.
\newblock In \emph{Proc. tenth international conference on machine learning},
  pp.\  97--103, 1993.

\bibitem[Feynman et~al.(1965)Feynman, Leighton, and Sands]{feynman1965feynman}
Richard~P Feynman, Robert~B Leighton, and Matthew Sands.
\newblock The feynman lectures on physics; vol. i.
\newblock \emph{American Journal of Physics}, 33\penalty0 (9):\penalty0
  750--752, 1965.

\bibitem[Gaucel et~al.(2014)Gaucel, Keijzer, Lutton, and
  Tonda]{gaucel2014learning}
S{\'e}bastien Gaucel, Maarten Keijzer, Evelyne Lutton, and Alberto Tonda.
\newblock Learning dynamical systems using standard symbolic regression.
\newblock In \emph{European Conference on Genetic Programming}, pp.\  25--36.
  Springer, 2014.

\bibitem[Greenwood et~al.(1986)Greenwood, Hanna, and Milton]{greenwood1986air}
Margaret~Stautberg Greenwood, Charles Hanna, and Rev John~W Milton.
\newblock Air resistance acting on a sphere: Numerical analysis, strobe
  photographs, and videotapes.
\newblock \emph{The Physics Teacher}, 24\penalty0 (3):\penalty0 153--159, 1986.

\bibitem[Hopcroft et~al.(2006)Hopcroft, Motwani, and
  Ullman]{hopcroft2006automata}
John~E Hopcroft, Rajeev Motwani, and Jeffrey~D Ullman.
\newblock Automata theory, languages, and computation.
\newblock \emph{International Edition}, 24\penalty0 (2), 2006.

\bibitem[Islam et~al.(2018)Islam, Kharma, and Grogono]{islam2018expansion}
Mohiul Islam, Nawwaf~N Kharma, and Peter Grogono.
\newblock Expansion: A novel mutation operator for genetic programming.
\newblock In \emph{IJCCI}, pp.\  55--66, 2018.

\bibitem[Kim et~al.(2019)Kim, Lu, Mukherjee, Gilbert, Jing, Ceperic, and
  Soljacic]{kim2019integration}
Samuel Kim, Peter Lu, Srijon Mukherjee, Michael Gilbert, Li~Jing, Vladimir
  Ceperic, and Marin Soljacic.
\newblock Integration of neural network-based symbolic regression in deep
  learning for scientific discovery.
\newblock \emph{arXiv preprint arXiv:1912.04825}, 2019.

\bibitem[Kocsis \& Szepesv{\'a}ri(2006)Kocsis and
  Szepesv{\'a}ri]{kocsis2006bandit}
Levente Kocsis and Csaba Szepesv{\'a}ri.
\newblock Bandit based monte-carlo planning.
\newblock In \emph{European conference on machine learning}, pp.\  282--293.
  Springer, 2006.

\bibitem[Koza \& Koza(1992)Koza and Koza]{koza1992genetic}
John~R Koza and John~R Koza.
\newblock \emph{Genetic programming: on the programming of computers by means
  of natural selection}, volume~1.
\newblock MIT press, 1992.

\bibitem[Kubal{\'\i}k et~al.(2019)Kubal{\'\i}k, {\v{Z}}egklitz, Derner, and
  Babu{\v{s}}ka]{kubalik2019symbolic}
Ji{\v{r}}{\'\i} Kubal{\'\i}k, Jan {\v{Z}}egklitz, Erik Derner, and Robert
  Babu{\v{s}}ka.
\newblock Symbolic regression methods for reinforcement learning.
\newblock \emph{arXiv preprint arXiv:1903.09688}, 2019.

\bibitem[Kusner et~al.(2017)Kusner, Paige, and
  Hern{\'a}ndez-Lobato]{kusner2017grammar}
Matt~J Kusner, Brooks Paige, and Jos{\'e}~Miguel Hern{\'a}ndez-Lobato.
\newblock Grammar variational autoencoder.
\newblock In \emph{Proceedings of the 34th International Conference on Machine
  Learning}, pp.\  1945--1954. JMLR. org, 2017.

\bibitem[Lample \& Charton(2019)Lample and Charton]{lample2019deep}
Guillaume Lample and Fran{\c{c}}ois Charton.
\newblock Deep learning for symbolic mathematics.
\newblock \emph{arXiv preprint arXiv:1912.01412}, 2019.

\bibitem[Langdon \& Gustafson(2010)Langdon and Gustafson]{langdon2010genetic}
William~B Langdon and Steven~M Gustafson.
\newblock Genetic programming and evolvable machines: ten years of reviews.
\newblock \emph{Genetic Programming and Evolvable Machines}, 11\penalty0
  (3):\penalty0 321--338, 2010.

\bibitem[Lindemuth(1971)]{lindemuth1971effect}
Jeffrey Lindemuth.
\newblock The effect of air resistance on falling balls.
\newblock \emph{American Journal of Physics}, 39\penalty0 (7):\penalty0
  757--759, 1971.

\bibitem[Long et~al.(2018)Long, Lu, Ma, and Dong]{long2018pde}
Zichao Long, Yiping Lu, Xianzhong Ma, and Bin Dong.
\newblock Pde-net: Learning pdes from data.
\newblock In \emph{International Conference on Machine Learning}, pp.\
  3208--3216. PMLR, 2018.

\bibitem[Long et~al.(2019)Long, Lu, and Dong]{long2019pde}
Zichao Long, Yiping Lu, and Bin Dong.
\newblock Pde-net 2.0: Learning pdes from data with a numeric-symbolic hybrid
  deep network.
\newblock \emph{Journal of Computational Physics}, 399:\penalty0 108925, 2019.

\bibitem[Lorenz(1963)]{lorenz1963deterministic}
Edward~N Lorenz.
\newblock Deterministic nonperiodic flow.
\newblock \emph{Journal of the atmospheric sciences}, 20\penalty0 (2):\penalty0
  130--141, 1963.

\bibitem[Lu et~al.(2021)Lu, Tao, Zhou, and Wang]{lu2021incorporating}
Qiang Lu, Fan Tao, Shuo Zhou, and Zhiguang Wang.
\newblock Incorporating actor-critic in monte carlo tree search for symbolic
  regression.
\newblock \emph{Neural Computing and Applications}, pp.\  1--17, 2021.

\bibitem[Ly \& Lipson(2012)Ly and Lipson]{ly2012learning}
Daniel~L Ly and Hod Lipson.
\newblock Learning symbolic representations of hybrid dynamical systems.
\newblock \emph{The Journal of Machine Learning Research}, 13\penalty0
  (1):\penalty0 3585--3618, 2012.

\bibitem[Martius \& Lampert(2017)Martius and Lampert]{martius2016extrapolation}
Georg~S Martius and Christoph Lampert.
\newblock Extrapolation and learning equations.
\newblock In \emph{5th International Conference on Learning Representations,
  ICLR 2017-Workshop Track Proceedings}, 2017.

\bibitem[Mundhenk et~al.(2021)Mundhenk, Landajuela, Glatt, Santiago, Faissol,
  and Petersen]{mundhenk2021symbolic}
T~Nathan Mundhenk, Mikel Landajuela, Ruben Glatt, Claudio~P Santiago, Daniel~M
  Faissol, and Brenden~K Petersen.
\newblock Symbolic regression via neural-guided genetic programming population
  seeding.
\newblock \emph{arXiv preprint arXiv:2111.00053}, 2021.

\bibitem[Petersen et~al.(2021)Petersen, Larma, Mundhenk, Santiago, Kim, and
  Kim]{petersen2020deep}
Brenden~K Petersen, Mikel~Landajuela Larma, Terrell~N Mundhenk, Claudio~Prata
  Santiago, Soo~Kyung Kim, and Joanne~Taery Kim.
\newblock Deep symbolic regression: Recovering mathematical expressions from
  data via risk-seeking policy gradients.
\newblock In \emph{International Conference on Learning Representations}, 2021.

\bibitem[Powell(1964)]{powell1964efficient}
Michael~JD Powell.
\newblock An efficient method for finding the minimum of a function of several
  variables without calculating derivatives.
\newblock \emph{The computer journal}, 7\penalty0 (2):\penalty0 155--162, 1964.

\bibitem[Quade et~al.(2016)Quade, Abel, Shafi, Niven, and
  Noack]{quade2016prediction}
Markus Quade, Markus Abel, Kamran Shafi, Robert~K Niven, and Bernd~R Noack.
\newblock Prediction of dynamical systems by symbolic regression.
\newblock \emph{Physical Review E}, 94\penalty0 (1):\penalty0 012214, 2016.

\bibitem[Rao et~al.(2022)Rao, Ren, Liu, and Sun]{rao2022discovering}
Chengping Rao, Pu~Ren, Yang Liu, and Hao Sun.
\newblock Discovering nonlinear {PDEs} from scarce data with physics-encoded
  learning.
\newblock In \emph{International Conference on Learning Representations}, pp.\
  1--19, 2022.

\bibitem[Rudy et~al.(2017)Rudy, Brunton, Proctor, and Kutz]{rudy2017data}
Samuel~H Rudy, Steven~L Brunton, Joshua~L Proctor, and J~Nathan Kutz.
\newblock Data-driven discovery of partial differential equations.
\newblock \emph{Science Advances}, 3\penalty0 (4):\penalty0 e1602614, 2017.

\bibitem[Sahoo et~al.(2018)Sahoo, Lampert, and Martius]{sahoo2018learning}
Subham Sahoo, Christoph Lampert, and Georg Martius.
\newblock Learning equations for extrapolation and control.
\newblock In \emph{International Conference on Machine Learning}, pp.\
  4442--4450, 2018.

\bibitem[Savitzky \& Golay(1964)Savitzky and Golay]{savitzky1964smoothing}
Abraham Savitzky and Marcel~JE Golay.
\newblock Smoothing and differentiation of data by simplified least squares
  procedures.
\newblock \emph{Analytical chemistry}, 36\penalty0 (8):\penalty0 1627--1639,
  1964.

\bibitem[Schmidt \& Lipson(2009)Schmidt and Lipson]{schmidt2009distilling}
Michael Schmidt and Hod Lipson.
\newblock Distilling free-form natural laws from experimental data.
\newblock \emph{science}, 324\penalty0 (5923):\penalty0 81--85, 2009.

\bibitem[Shah et~al.(2019)Shah, Xie, and Xu]{shah2019non}
Devavrat Shah, Qiaomin Xie, and Zhi Xu.
\newblock Non-asymptotic analysis of monte carlo tree search.
\newblock \emph{arXiv preprint arXiv:1902.05213}, 2019.

\bibitem[Shampine(1975)]{shampine1975computer}
Lawrence~F Shampine.
\newblock Computer solution of ordinary differential equations.
\newblock \emph{The initial value problem}, 1975.

\bibitem[Shampine \& Reichelt(1997)Shampine and Reichelt]{shampine1997matlab}
Lawrence~F Shampine and Mark~W Reichelt.
\newblock The matlab ode suite.
\newblock \emph{SIAM journal on scientific computing}, 18\penalty0
  (1):\penalty0 1--22, 1997.

\bibitem[Silver et~al.(2017)Silver, Schrittwieser, Simonyan, Antonoglou, Huang,
  Guez, Hubert, Baker, Lai, Bolton, et~al.]{silver2017mastering}
David Silver, Julian Schrittwieser, Karen Simonyan, Ioannis Antonoglou, Aja
  Huang, Arthur Guez, Thomas Hubert, Lucas Baker, Matthew Lai, Adrian Bolton,
  et~al.
\newblock Mastering the game of go without human knowledge.
\newblock \emph{Nature}, 550\penalty0 (7676):\penalty0 354--359, 2017.

\bibitem[Sun et~al.(2021)Sun, Liu, and Sun]{sun2021physics}
Fangzheng Sun, Yang Liu, and Hao Sun.
\newblock Physics-informed spline learning for nonlinear dynamics discovery.
\newblock In \emph{Proceedings of the Thirtieth International Joint Conference
  on Artificial Intelligence}, pp.\  2054--2061, 2021.

\bibitem[Udrescu \& Tegmark(2020)Udrescu and Tegmark]{udrescu2020ai}
Silviu-Marian Udrescu and Max Tegmark.
\newblock Ai feynman: A physics-inspired method for symbolic regression.
\newblock \emph{Science Advances}, 6\penalty0 (16):\penalty0 eaay2631, 2020.

\bibitem[Udrescu \& Tegmark(2021)Udrescu and Tegmark]{udrescu2021symbolic}
Silviu-Marian Udrescu and Max Tegmark.
\newblock Symbolic pregression: discovering physical laws from distorted video.
\newblock \emph{Physical Review E}, 103\penalty0 (4):\penalty0 043307, 2021.

\bibitem[Udrescu et~al.(2020)Udrescu, Tan, Feng, Neto, Wu, and
  Tegmark]{udrescu2020ai2}
Silviu-Marian Udrescu, Andrew Tan, Jiahai Feng, Orisvaldo Neto, Tailin Wu, and
  Max Tegmark.
\newblock Ai feynman 2.0: Pareto-optimal symbolic regression exploiting graph
  modularity.
\newblock \emph{arXiv preprint arXiv:2006.10782}, 2020.

\bibitem[Uy et~al.(2011)Uy, Hoai, O’Neill, McKay, and
  Galv{\'a}n-L{\'o}pez]{uy2011semantically}
Nguyen~Quang Uy, Nguyen~Xuan Hoai, Michael O’Neill, Robert~I McKay, and Edgar
  Galv{\'a}n-L{\'o}pez.
\newblock Semantically-based crossover in genetic programming: application to
  real-valued symbolic regression.
\newblock \emph{Genetic Programming and Evolvable Machines}, 12\penalty0
  (2):\penalty0 91--119, 2011.

\bibitem[Vaddireddy et~al.(2020)Vaddireddy, Rasheed, Staples, and
  San]{vaddireddy2020feature}
Harsha Vaddireddy, Adil Rasheed, Anne~E Staples, and Omer San.
\newblock Feature engineering and symbolic regression methods for detecting
  hidden physics from sparse sensor observation data.
\newblock \emph{Physics of Fluids}, 32\penalty0 (1):\penalty0 015113, 2020.

\bibitem[White et~al.(2015)White, Yoo, and Singer]{white2015programming}
David~R White, Shin Yoo, and Jeremy Singer.
\newblock The programming game: evaluating mcts as an alternative to gp for
  symbolic regression.
\newblock In \emph{Proceedings of the Companion Publication of the 2015 Annual
  Conference on Genetic and Evolutionary Computation}, pp.\  1521--1522, 2015.

\end{thebibliography}
\bibliographystyle{iclr2023_conference}

\clearpage

\appendix
\counterwithin{figure}{section}
\counterwithin{table}{section}

\section*{Appendix}


\section{\hsedit{Hyperparameter Setting}}\label{appendix:hyperparameter}

\hsedit{We perform a parametric study on the value of discount factor $\eta$ based on Nguyen's benchmark problems without measurement noise. Empirically, setting $\eta=0.9999$ ensures the scores of ground truth equations stand out and successfully enforces the sparsity in all experiments. For the discovery of very chaotic dynamical systems based on measurement data, we expect some physics terms from the governing equations to have a weak impact on the state variables (e.g., in the chaotic double pendulum system experiments, the effects from physics terms $\sin(\theta_1)$ and $\sin(\theta_2)$ are hard to be captured), and the effect of data noise is unknown. Hence, we set $\eta=1$ to leverage the full strength of data fitting to enable the detection of physics terms that are offset or overwhelmed by data noise but are pivotal to the systems. Nevertheless, we must acknowledge that this selection process is empirical, which depends on our desire for the degree of parsimony of the target equation(s).

As for the hyperparameters in the training schema (i.e., maximum module transplantation, episodes, maximum tree size, maximum augmented grammars), we have conducted parametric convergence tests for each experiment to ensure the learning curves (i.e., maximum scores in the history) converge. For example, as discussed in Section \ref{appendix:nguyen}, Table \ref{Table:Nguyen Parameter} shows the setting of these hyperparameters.}

\section{Nguyen's Benchmark Problems}\label{appendix:nguyen}

This section provides more detailed experiment settings for \hsedit{the} Nguyen's benchmark tasks that are described in Section \ref{ss:SR1} of the main \hsedit{text, where} the training hyperparameters for the SPL machine in these equation discovery experiments \hsedit{are also listed}. Table \ref{Table:Nguyen Operators} presents the candidate mathematical operations allowed for three tested models and \hsedit{Table \ref{Table:Nguyen Parameter} displays training hyperparameters for the SPL machine in the Nguyen's benchmark tasks.}

\begin{table}[b!]
\centering
\caption{Candidate operators for each Nguyen's benchmark task. $const$ denotes constant values. }
\vspace{6pt}
{
\begin{tabular}{ll}
\toprule
\textbf{Benchmark}  &  \textbf{Candidate Operations}\\
    \midrule
    Nguyen-1 & $+,-,\times,\div,\cos(\cdot),\sin(\cdot),\exp(\cdot)$\\
    Nguyen-2 & $+,-,\times,\div,\cos(\cdot),\sin(\cdot),\exp(\cdot)$ \\
    Nguyen-3 & $+,-,\times,\div,\cos(\cdot),\sin(\cdot),\exp(\cdot)$ \\
    Nguyen-4 & $+,-,\times,\div,\cos(\cdot),\sin(\cdot),\exp(\cdot)$ \\
    Nguyen-5 & $+,-,\times,\div,\cos(\cdot),\sin(\cdot),\exp(\cdot)$ \\
    Nguyen-6 & $+,-,\times,\div,\cos(\cdot),\sin(\cdot),\exp(\cdot)$ \\
    Nguyen-7 & $+,-,\times,\div,\cos(\cdot),\sin(\cdot),\exp(\cdot), \log(\cdot), \sqrt{\cdot}$ \\
    Nguyen-8 & $+,-,\times,\div,\cos(\cdot),\sin(\cdot),\exp(\cdot), \log(\cdot), \sqrt{\cdot}$ \\
    Nguyen-9 & $+,-,\times,\div,\cos(\cdot),\sin(\cdot),\exp(\cdot)$ \\
    Nguyen-10 & $+,-,\times,\div,\cos(\cdot),\sin(\cdot),\exp(\cdot)$ \\
    Nguyen-11 & $+,-,\times,\div,\cos(\cdot),\sin(\cdot),\exp(\cdot), \log(\cdot), \sqrt{\cdot}$ \\
    Nguyen-12 & $+,-,\times,\div,\cos(\cdot),\sin(\cdot),\exp(\cdot)$ \\
    Nguyen-1$^c$ & $+,-,\times,\div,\cos(\cdot),\sin(\cdot),\exp(\cdot), const$ \\
    Nguyen-2$^c$ & $+,-,\times,\div,\cos(\cdot),\sin(\cdot),\exp(\cdot), const$ \\
    Nguyen-5$^c$ & $+,-,\times,\div,\cos(\cdot),\sin(\cdot),\exp(\cdot), const$ \\
    Nguyen-8$^c$ & $+,-,\times,\div,\cos(\cdot),\sin(\cdot),\exp(\cdot), \log(\cdot), \sqrt{\cdot}, const$ \\
    Nguyen-9$^c$ & $+,-,\times,\div,\cos(\cdot),\sin(\cdot),\exp(\cdot), const$ \\
    \bottomrule
\end{tabular}
}
\label{Table:Nguyen Operators}
\end{table}

\begin{table}[t!]
\centering
\caption{Training Hyperparameter settings for the SPL model in Nguyen's benchmark tasks. }
\vspace{6pt}
{
\begin{tabular}{lllll}
\toprule
\scriptsize\textbf{Benchmark}  &  \scriptsize\textbf{Maximum Module} & \scriptsize\textbf{Episodes Between} & \scriptsize\textbf{Maximum} & \scriptsize\textbf{Maximum Augmented}\\
&  \scriptsize\textbf{Transplantation}  & \scriptsize\textbf{Module Transplantation} & \scriptsize\textbf{Tree Size} &  \scriptsize\textbf{Grammars}\\
    \midrule
    Nguyen-1 & 20 & 10,000 & 50 & 5 \\
    Nguyen-2 & 20 & 10,000 & 50 & 5 \\
    Nguyen-3 & 20 & 100,000 & 50 & 5 \\
    Nguyen-4 & 20 & 100,000 & 50 & 5 \\
    Nguyen-5 & 20 & 100,000 & 50 & 5 \\
    Nguyen-6 & 20 & 10,000 & 50 & 5 \\
    Nguyen-7 & 20 & 5,000 & 50 & 5 \\
    Nguyen-8 & 20 & 5,000 & 50 & 5 \\
    Nguyen-9 & 20 & 10,000 & 50 & 5 \\
    Nguyen-10 & 20 & 10,000 & 50 & 5 \\
    Nguyen-11 & 20 & 10,000 & 50 & 5 \\
    Nguyen-12 & 20 & 100,000 & 50 & 5 \\
    Nguyen-1$^c$ & 20 & 2,000 & 50 & 5 \\
    Nguyen-2$^c$ & 20 & 10,000 & 50 & 5 \\
    Nguyen-5$^c$ & 20 & 10,000 & 50 & 5 \\
    Nguyen-8$^c$ & 20 & 2,000 & 50 & 5 \\
    Nguyen-9$^c$ & 20 & 1,000 & 50 & 5 \\
    \bottomrule
\end{tabular}
}
\label{Table:Nguyen Parameter}
\end{table}

Moreover, the utilization of CFG in the SPL machine facilitates the flexibility of applying some prior knowledge including the universal mathematical rules and constraints. This feature empirically turns out to be an accessible and scalable approach for avoiding meaningless mathematical expressions. In the Nguyen's benchmark tasks, one or multiple mathematical constraints are given to the SPL machine. These constraints include 
\begin{enumerate}
    \item Only variables and constant values \hsedit{are} allowed in trigonometric functions. \vspace{-1pt}
    \item Variables in trigonometric functions, logarithms and roots are up to the polynomial of 3. \vspace{-1pt}
    \item Trigonometric functions, logarithms and roots are not allowed to form unreasonable composite functions with each other, such as $\sin(\cos(...))$. \vspace{-1pt}
    \item Some small integers (e.g. 1, 2) are used directly as leaves. 
\end{enumerate}
They can be easily implemented into the SPL machine by defining \hsedit{or} adjusting non-terminal nodes or production rules in the customized CFG. Constraints adopted by each task \hsedit{are} shown in Table \ref{Table:Nguyen Constraints}.

\begin{table}[t!]
\centering
\vspace{12pt}
\caption{Mathematical constraints for the SPL machine in each Nguyen's benchmark problem. }
\vspace{6pt}
{
\begin{tabular}{ll}
\toprule
\textbf{Benchmark}  &  \textbf{Constraints Utilization}\\
    \midrule
    Nguyen-1 & \{1, 3\} \\
    Nguyen-2 & \{1, 3\} \\
    Nguyen-3 & \{1, 3\} \\
    Nguyen-4 & \{1, 3\} \\
    Nguyen-5 & \{2, 3\} \\
    Nguyen-6 & \{2, 3\} \\
    Nguyen-7 & \{2, 3\} \\
    Nguyen-8 & \{\} \\
    Nguyen-9 & \{2, 3, 4\} \\
    Nguyen-10 & \{2, 3, 4\} \\
    Nguyen-11 & \{2, 3\} \\
    Nguyen-12 & \{1, 3, 4\} \\
    Nguyen-1$^c$ & \{1, 3\} \\
    Nguyen-2$^c$ & \{1, 3\} \\
    Nguyen-5$^c$ & \{2, 3\} \\
    Nguyen-8$^c$ & \{\} \\
    Nguyen-9$^c$ & \{2, 3, 4\} \\
    \bottomrule
\end{tabular}
}
\label{Table:Nguyen Constraints}
\vspace{6pt}
\end{table}

\section{\hsedit{Results of Ablation Study}} \label{appendix:ablation}
\hsedit{We consider four ablation studies by removing the following:
\begin{enumerate}[(a)]
    \item the adaptive scaling in reward calculation, \vspace{-2pt}
    \item the discount factor $\eta^n$ that drives equation parsimony in Eq. \eref{eq:reward}, \vspace{-2pt}
    \item module transplantation in tree generation, \vspace{-2pt}
    \item all of the above three.
\end{enumerate}
The resulting models are denoted with Model A, Model B, Model C, and Model D (note that Model D is equivalent to the vanilla MCTS). The ablation study is performed on the 12 classic Nguyen's benchmark problems, where the recovery rate of each model is calculated. 
The results of the ablation study are summarized in Table \ref{Table:Appendix_ablation}. It is clear that module transplantation brings the largest gain in recovery rate. All the ablated models fail to uncover the Nguyen-12 equation (the most difficult case), where the coefficient 1/2 needs to be represented by mathematical operators and symbols, e.g., $x/(x + x)$, $y/(y + y)$, etc.}

\begin{table}[b!]\color{black}
    \centering
    \caption{Summary of the ablation study results.}
    \vspace{6pt}
    \small
    \begin{tabular}{ccccccc}
    \toprule
    \textbf{Benchmark} & \textbf{Expression}  &  \textbf{SPL} &  \textbf{Model A} & \textbf{Model B} & \textbf{Model C} & \textbf{Model D} \\
    \midrule
    Nguyen-1 & $x^3+x^2+x$                      & 100\% & 100\% & 100\% & 14\% & 12\%  \\ 
    Nguyen-2 & $x^4+x^3+x^2+x$                  & 100\% & 100\% & 100\% & 0\% & 0\%  \\
    Nguyen-3 & $x^5+x^4+x^3+x^2+x$ 	            & 100\% & 100\% & 100\% & 0\% & 0\%  \\
    Nguyen-4 & $x^6+x^5+x^4+x^3+x^2+x$          & 99\% & 96\% & 92\% & 0\% & 0\%  \\
    Nguyen-5 & $\sin(x^2)\cos(x)-1$             & 95\% & 95\% & 92\% & 92\% & 88\%  \\
    Nguyen-6 & $\sin(x^2)+\sin(x+x^2)$	        & 100\% & 100\% & 96\% & 100\% & 100\%  \\
    Nguyen-7 & $\ln(x+1)+\ln(x^2+1)$            & 100\% & 100\% & 100\% & 100\% & 100\%  \\
    Nguyen-8 & $\sqrt{x}$                       & 100\% & 100\% & 100\% & 100\% & 100\%  \\
    Nguyen-9 & $\sin(x)+\sin(y^2)$ 	            & 100\% & 100\% & 100\% & 100\% & 100\%  \\
    Nguyen-10 & $2\sin(x)\cos(y)$               & 100\% & 100\% & 100\% & 0\% & 0\%  \\
    Nguyen-11 & $x^y$                           & 100\% & 96\% & 96\% & 92\% & 87\%  \\
    Nguyen-12 & $x^4-x^3+\frac{1}{2}y^2-y$	    & 28\% & 0\% & 0\% & 0\% & 0\%  \\
    \midrule
    Average &                                   & \textbf{93.5\%} & 90.58\% & 89.67\% & 49.83\% & 48.92\%  \\    
    \bottomrule
    \end{tabular}
    \label{Table:Appendix_ablation}
    \vspace{12pt}
    \normalsize
\end{table}

\section{Free Falling Balls with Air Resistance} \label{appendix:free falling}

\begin{figure}[t!]
	\centering
	\includegraphics[width=0.95\linewidth]{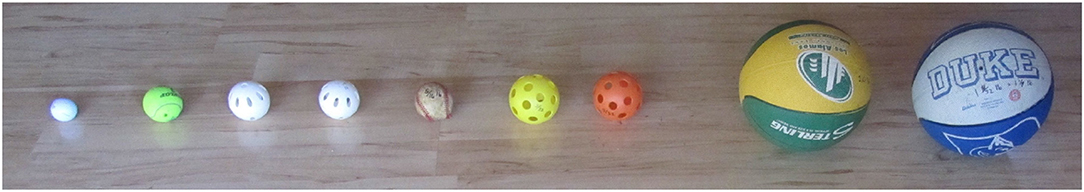} 
	\caption{The experimental balls that were dropped from the bridge \citep{de2020discovery}. From left to right: golf ball, tennis ball, whiffle ball 1, whiffle ball 2, baseball, yellow whiffle ball, orange whiffle ball, green basketball, and blue basketball. Volleyball is not shown here.} 
	\label{Figure:balls}
\end{figure}

This appendix section reveals more details \hsedit{on discovering} the \hsedit{physical} laws, \hsedit{in the context} of relationships between height and time, \hsedit{for} the cases of free-falling objects with air resistance based on multiple experimental ball-drop datasets \citep{de2020discovery}. The datasets contain the records of 11 different types of balls, as shown in Figure \ref{Figure:balls}, dropped from a bridge, \hsedit{collected at} a 30 Hz sampling rate. The time between dropping and landing varies in each case due to the fact that air resistance has different effects on these free-dropping balls and induces divergent \hsedit{physical} laws. Consequently, we consider each ball as an individual experiment and discover \hsedit{the physical law} for each of them. The measurement dataset of a free-dropping ball is split into a training set (records from the first 2 seconds, 60 \hsedit{measurements}) and a testing set (records after 2 seconds). 

\begin{figure}[t!]
	\centering
	\includegraphics[width=0.97\linewidth]{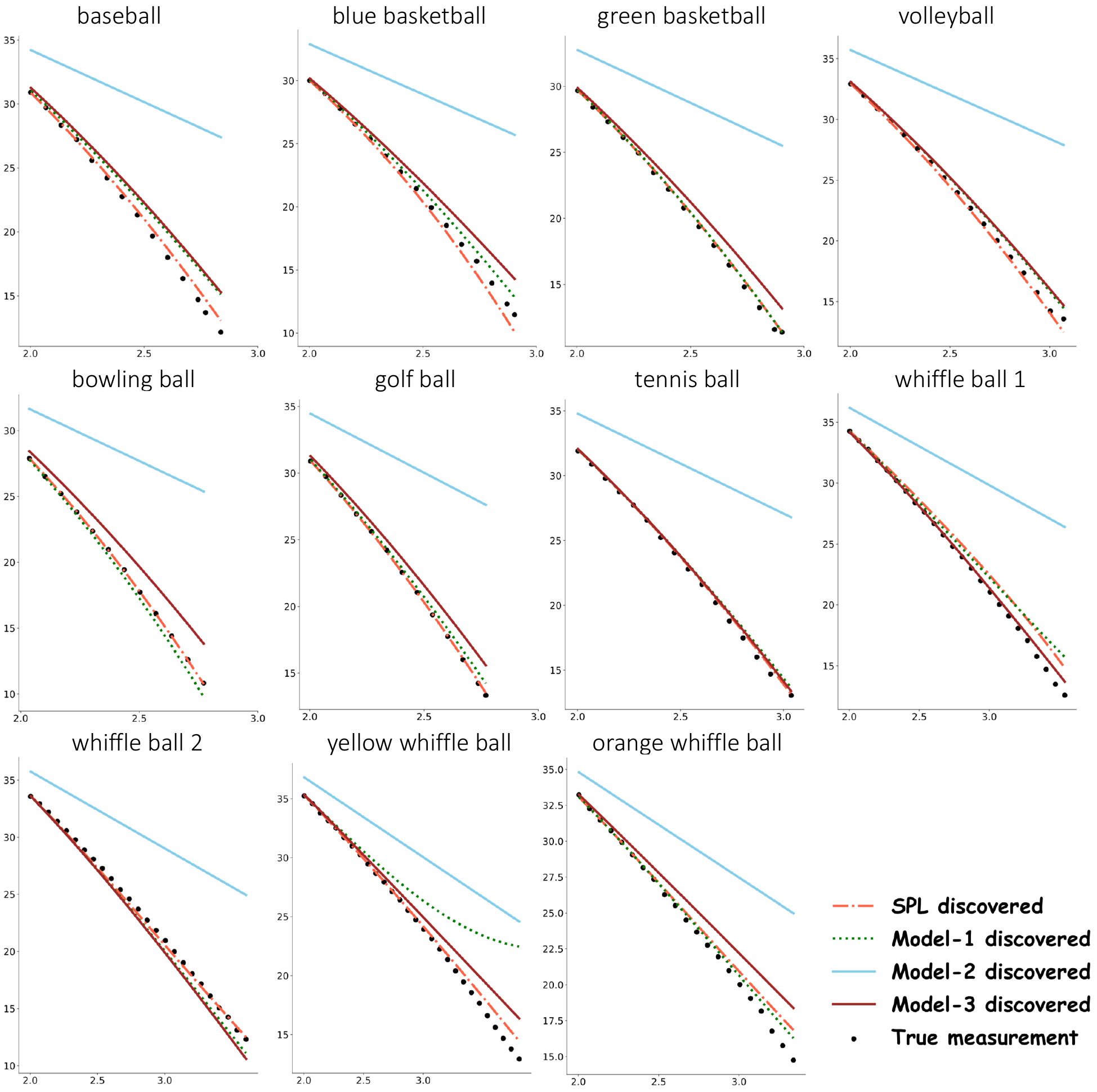}
	\caption{Trajectories after 2 seconds predicted by uncovered \hsedit{physical} laws.} 
	\label{Figure:ball pred}
\end{figure}

The \hsedit{physical} laws \hsedit{of} the three \hsedit{baseline} models and \hsedit{distilled by} the SPL machine from training data are exhibited in Table \ref{Table:ball eq}. These discovered \hsedit{physical laws} are then applied to forecast the height of the balls at the time slots in \hsedit{the} testing dataset. These predictions, in comparison with the ground truth trajectory recorded, are shown in Figure \ref{Figure:ball pred}. \hsedit{The} prediction error is shown in the Table \ref{Table:ball mse} of the main \hsedit{text}.

\begin{table}[htbp]
    \centering
    \caption{Uncovered physics from the motions of \hsedit{free-falling} balls by the SPL machine and three baseline models. \hsedit{Note that these formulas are the raw equations produced by SRL. Further simplification helps better parsimony of the formulas.}}
    \vspace{6pt}
    \small
    \begin{tabular}{lll}
    \toprule
    \textbf{Type} &  \textbf{Model} &  \textbf{Expression} \\
    \midrule
    baseball    & SPL     & \scriptsize$H(t)=47.8042+0.6253t-4.5383t^2$ \\[-0.3ex]
                & Model-1 & \scriptsize$H(t)=47.682+1.456t-5.629t^2+0.376t^3$ \\[-0.3ex]
                & Model-2 & \scriptsize$H(t)=45.089-8.156t+5.448\exp(0t)$ \\[-0.3ex]
                & Model-3 & \scriptsize$H(t)=48.051-183.467\log(\cosh(0.217t))$\\[-0.3ex]
    \midrule
    blue        & SPL     & \scriptsize$H(t)=46.4726-5.105t^2+t^3-0.251t^4$ \\[-0.3ex]
    basketball  & Model-1 & \scriptsize$H(t)=46.513-0.493t-3.912t^2+0.03t^3$ \\[-0.3ex]
                & Model-2 & \scriptsize$H(t)=43.522-7.963t+5.306\exp(0t)$ \\[-0.3ex]
                & Model-3 & \scriptsize$H(t)=46.402-84.791\log(\cosh(0.319t))$\\[-0.3ex]
    \midrule
    green       & SPL     & \scriptsize$H(t)=45.9087-4.1465t^2+\log(\cosh(1))$ \\[-0.3ex]
    basketball  & Model-1 & \scriptsize$H(t)=46.438-0.34t-3.882t^2-0.055t^3$ \\[-0.3ex]
                & Model-2 & \scriptsize$H(t)=43.512-8.043t+5.346\exp(0t)$ \\[-0.3ex]
                & Model-3 & \scriptsize$H(t)=46.391-124.424\log(\cosh(0.263t))$\\[-0.3ex]
    \midrule
    volleyball  & SPL     & \scriptsize$H(t)=48.0744-3.7772t^2$\\[-0.3ex]
                & Model-1 & \scriptsize$H(t)=48.046+0.362t-4.352t^2+0.218t^3$\\[-0.3ex]
                & Model-2 & \scriptsize$H(t)=45.32-7.317t+5.037\exp(0t)$\\[-0.3ex]
                & Model-3 & \scriptsize$H(t)=48.124-107.816\log(\cosh(0.27t))$\\[-0.3ex]
    \midrule
    bowling     & SPL     &
\scriptsize$H(t)=46.1329-3.8173t^2-0.2846t^3+4.14\times10^{-5}\exp(20.7385t^2)\exp(-12.4538t^3)$\\[-0.3ex]
    ball        & Model-1 & \scriptsize$H(t)=46.139-0.091t-3.504t^2-0.431t^3$\\[-0.3ex]
                & Model-2 & \scriptsize$H(t)=43.336-8.525t+5.676\exp(0t)$\\[-0.3ex]
                & Model-3 & \scriptsize$H(t)=46.342-247.571\log(\cosh(0.189t))$\\[-0.3ex]
    \midrule
    golf ball   & SPL     & \scriptsize$H(t)=49.5087-4.9633t^2+\log(\cosh(t))$\\[-0.3ex]
                & Model-1 & \scriptsize$H(t)=49.413+0.532t-5.061t^2+0.102t^3$\\[-0.3ex]
                & Model-2 & \scriptsize$H(t)=46.356-8.918t+5.964\exp(0t)$\\[-0.3ex]
                & Model-3 & \scriptsize$H(t)=49.585-178.47\log(\cosh(0.23t))$\\[-0.3ex]
    \midrule
    tennis      & SPL     & \scriptsize$H(t)=47.8577-4.0574t^2+\log(\cosh(0.121t^3))$\\[-0.3ex]
    ball        & Model-1 & \scriptsize$H(t)=47.738+0.658t-4.901t^2+0.325t^3$\\[-0.3ex]
                & Model-2 & \scriptsize$H(t)=45.016-7.717t+5.212\exp(0t)$\\[-0.3ex]
                & Model-3 & \scriptsize$H(t)=47.874-114.19\log(\cosh(0.269t))$\\[-0.3ex]
    \midrule
    whiffle     & SPL     & \scriptsize$H(t)=4.1563t^2-t^3+47.0133\exp(-0.1511t^2)$\\[-0.3ex]
    ball        & Model-1 & \scriptsize$H(t)=46.969+0.574t-4.505t^2+0.522t^3$\\[-0.3ex]
    1           & Model-2 & \scriptsize$H(t)=44.259-6.373t+4.689\exp(0t)$\\[-0.3ex]
                & Model-3 & \scriptsize$H(t)=47.062-34.083\log(\cosh(0.462t))$\\[-0.3ex]
    \midrule
    whiffle     & SPL     & \scriptsize$H(t)=-18.6063+65.8583\exp(-0.0577t^2)$\\[-0.3ex]
    ball        & Model-1 & \scriptsize$H(t)=47.215+0.296t-4.379t^2+0.421t^3$\\[-0.3ex]
    2           & Model-2 & \scriptsize$H(t)=44.443-6.744t+4.813\exp(0t)$\\[-0.3ex]
                & Model-3 & \scriptsize$H(t)=47.255-38.29\log(\cosh(0.447t))$\\[-0.3ex]
    \midrule
    yellow      & SPL     & \scriptsize$H(t)=148.9911/(\log(\cosh(t))+3.065)-14.5828t^2/(\log(\cosh(t))+3.065)$\\[-0.3ex]
    whiffle     &         & \scriptsize$\quad\quad\quad+48.6092\log(\cosh(t))/(\log(\cosh(t))+3.065)$\\[-0.3ex]
    ball        & Model-1 & \scriptsize$H(t)=48.613-0.047t-4.936t^2+0.826t^3$\\[-0.3ex]
                & Model-2 & \scriptsize$H(t)=45.443-6.789t+4.973\exp(0t)$\\[-0.3ex]
                & Model-3 & \scriptsize$H(t)=48.594-12.49\log(\cosh(0.86t))$\\[-0.3ex]
    \midrule
    orange      & SPL     & \scriptsize$H(t)=-1.6626t+47.8622\exp(-0.06815t^2)$\\[-0.3ex]
    whiffle     & Model-1 & \scriptsize$H(t)=47.836-1.397t-3.822t^2+0.422t^3$\\[-0.3ex]
    ball        & Model-2 & \scriptsize$H(t)=44.389-7.358t+5.152\exp(0t)$\\[-0.3ex]
                & Model-3 & \scriptsize$H(t)=47.577-12.711\log(\cosh(0.895t))$\\[-0.3ex]
    \bottomrule
    \end{tabular}
    \label{Table:ball eq}
    \normalsize
\end{table}

\section{Nonlinear Dynamics Discovery} \label{appendix:nonlinear dynamics}

This appendix section \hsedit{elaborates more details on} the two governing equation discovery experiments presented in the main \hsedit{text}, including the datasets and full results. 

\subsection{Lorenz System}

\begin{figure}[t!]
	\centering
	\includegraphics[width=0.85\linewidth]{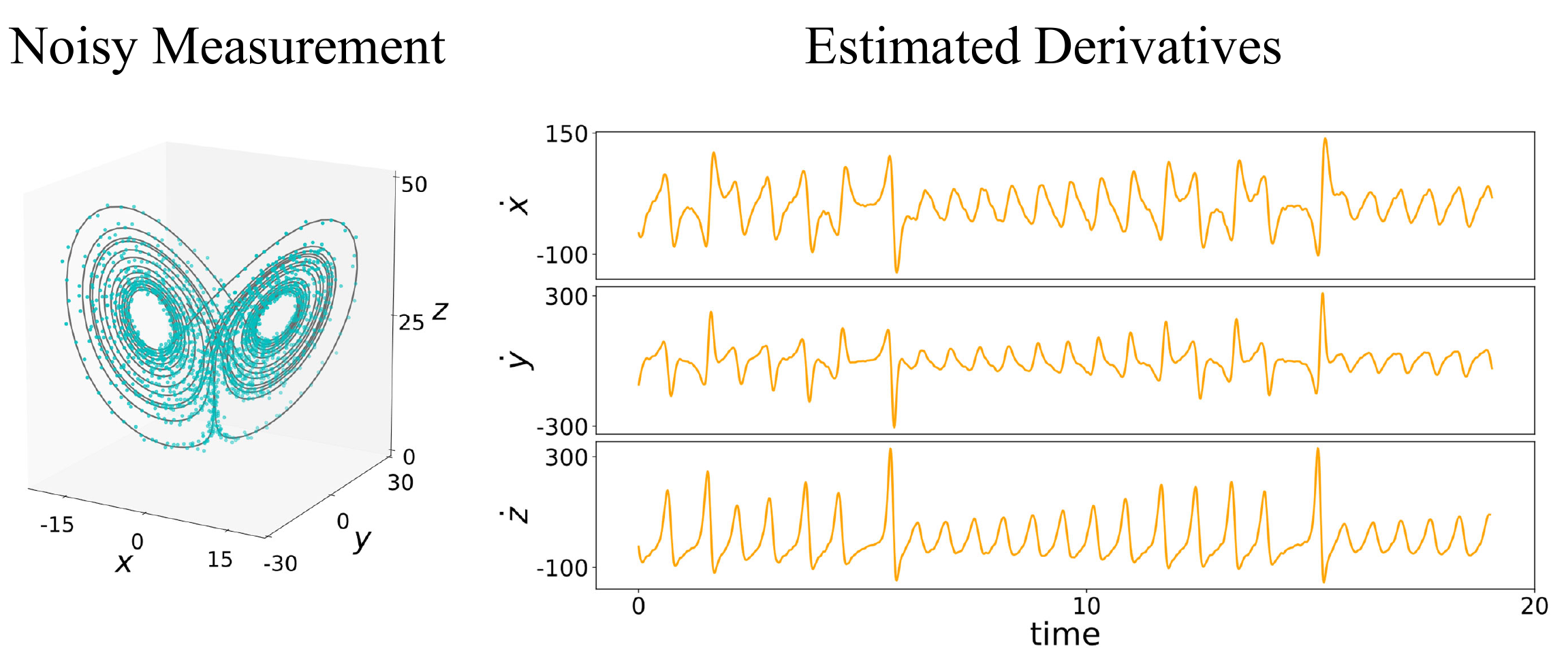}
	\caption{Lorenz system for \hsedit{the} experiment. Noisy measurement data and numerically estimated derivatives smoothed by Savitzky–Golay filter.} 
	\label{Figure:Lorenz data}
\end{figure}

The 3-dimensional Lorenz system is governed by 
\begin{equation}\label{Eq:Lorenz}
\begin{split}
    \dot{x}& = \sigma(y-x)\\
    \dot{y}& = x(\rho-z)-y\\
    \dot{z}& = xy-\beta z
\end{split}
\end{equation}
with parameters $\sigma=10$, $\beta=8/3$, and $\rho=28$, under which the Lorenz attractor has two lobes and the system, starting from anywhere, makes cycles around one lobe before switching to the other and iterates repeatedly. The measurement data of \hsedit{the} Lorenz system \hsedit{states} in this experiment contains a clean signal with 5\% Gaussian white noise. The derivatives of Lorenz's state variables are \hsedit{unmeasured but} numerically estimated and smoothed by the Savitzky–Golay filter. The noisy synthetic measurement \hsedit{data} and numerically obtained derivatives are shown in Figure \ref{Figure:Lorenz data}.

\begin{table}[t!]
\centering
\caption{Discovered governing equations for the Lorenz system.}
\vspace{6pt}
{\small
\begin{tabular}{ll}
\toprule
\textbf{Model}  &  \textbf{Discovered Governing Equations} \\
\midrule
\small Eureqa     & \small$\dot{x} = -0.56 - 9.02x + 9.01y$ \\
                  & \small$\dot{y} = -0.047 + 18.79x + 1.86y - 0.046xy -0.74xz$ \\
                  & \small$\dot{z} = -3.04 - 2.23z + 0.88xy$\\
\midrule
\small pySINDy    & \small$\dot{x} = -0.46-9.18x+9.17y$ \\
                  & \small$\dot{y} = 22.32x+0.15y-0.85xz$ \\
                  & \small$\dot{z} = 6.04 - 2.83z + 0.15x^2 + 0.81xy$\\
\midrule
\small NGGP       & \small$\dot{x} = 0.0047 - 10.02x + 10.01y - 0.007x^2 + 0.007xy - 0.37x/z - 0.00074x^2y$\\
                  & \small$\qquad + \, 0.00063x^3 + 0.00018x^2z + 0.00011xy^2 - 6.59e^{-5}xyz - 0.00011y^z$\\
                  & \small$\dot{y} = 26.36x - 1.5y - 0.83xz - 7.20x/z + 13.08y/z + 4.52e^{-5}x^3 - 0.0038xz^2$\\
                  & \small$\qquad - \, 44.25x/z^2 + 0.00028x^3/z - 0.00017x^3z + 5.98e^{-6}x^3z^2$\\
                  & \small$\dot{z} = - 0.64 - 0.036y - 2.64z + 1.038xy + 0.00021xz + 0.0011yz - 0.00021x/z $\\
                  & \small$\qquad - \, 8.04e^{-8}y/x + 0.00022y/z - 8.04e^{-8}z^2 -0.00021xyz + 0.00021xy/z$\\
                  & \small$\qquad - \, 0.001y^2z - 0.00021y^2/z + 1.17y/z^2 - 3.89e^{-7}yz^2/x + 6.66e^{-9}z^2/x$\\

\midrule
SPL             & $\dot{x} = -9.966x + 9.964y$\\
                & $\dot{y} = 27.764x - 0.942y - 0.994xz$ \\
                & $\dot{z} = -2.655z + 0.996xy$\\
\midrule
\cellcolor{mygray}True  & \cellcolor{mygray}$\dot{x} = -10x + 10y$\\
\cellcolor{mygray}      & \cellcolor{mygray}$\dot{y} = 28x-y-xz$\\
\cellcolor{mygray}      & \cellcolor{mygray}$\dot{z} = -2.667z+xy$\\
\bottomrule
\end{tabular}
}
\label{Table:Lorenz}
\vspace{6pt}
\end{table}

\begin{figure}[t!]
	\centering
	\includegraphics[width=0.75\linewidth]{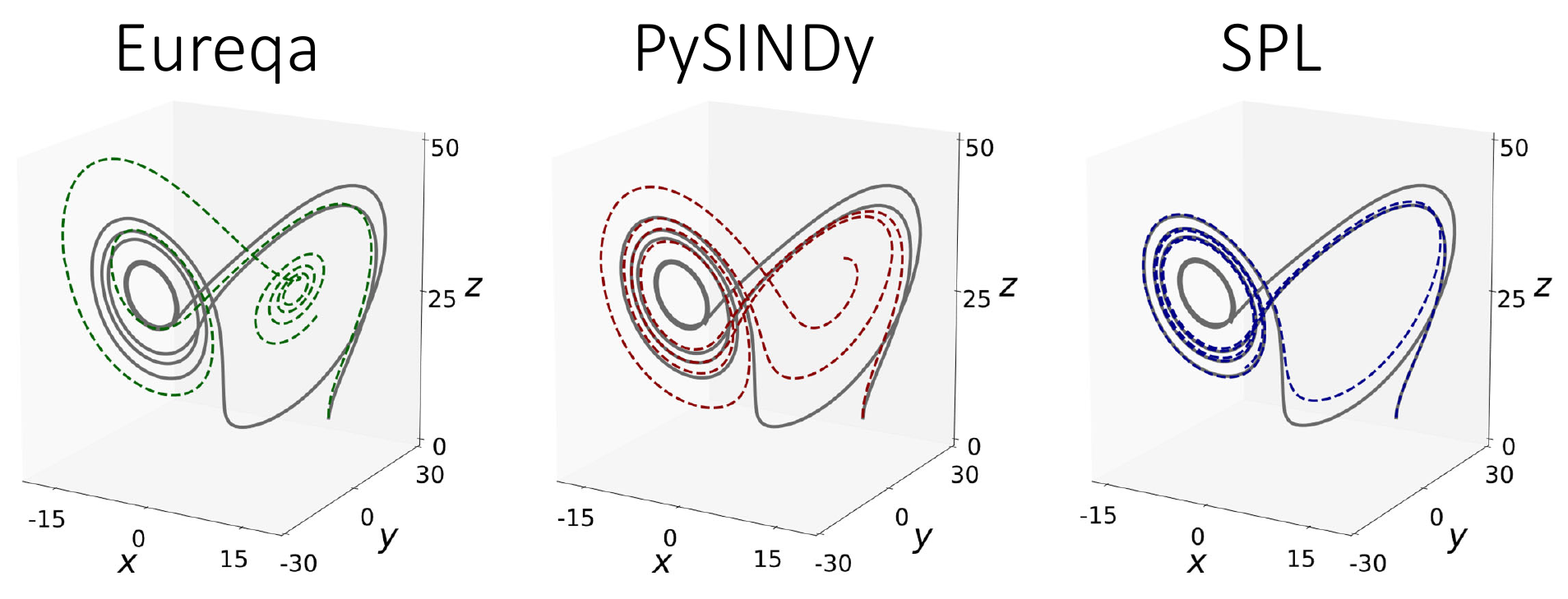}
	\caption{Response prediction for 5 seconds by identified governing equations (dashed plots) under a different validation IC of Lorenz system, in comparison with the ground truth trajectory (grey).} 
	\label{Figure:Lorenz pred}
\end{figure}

Table \ref{Table:Lorenz} presents the distilled governing equations \hsedit{for} the Lorenz system by \hsedit{the SPL machine compared with 3 baseline} models. It is observed that the SPL \hsedit{machine uncovers} the explicit form of equations accurately in the context of active terms, whereas Eureqa, pySINDy and NGGP yield several false-positive terms in \hsedit{the underlying} governing equations. \hsedit{The} predicted system responses (starting from a different initial condition) simulated from these uncovered equations are shown in Figure \ref{Figure:Lorenz pred}. \hsedit{Although} it is hard to reproduce the most accurate coefficients due to the tremendous errors \hsedit{induced by} numerical differentiation of noisy measurement \hsedit{data} as depicted in Figure \ref{Figure:Lorenz data}, the SPL machine is still capable of distilling the most concise symbolic combination of operators and variables to correctly formulate \hsedit{the} parsimonious mathematical expressions that govern the Lorenz system \hsedit{dynamics}. The predicted responses \hsedit{for} the governing equations unearthed by the SPL machine \hsedit{simulate} the system in a decent manner.

\subsection{Mounted Double Pendulum System}\label{appen:DoublePendulum}
\begin{figure}[t!]
	\centering
	\includegraphics[width=0.99\linewidth]{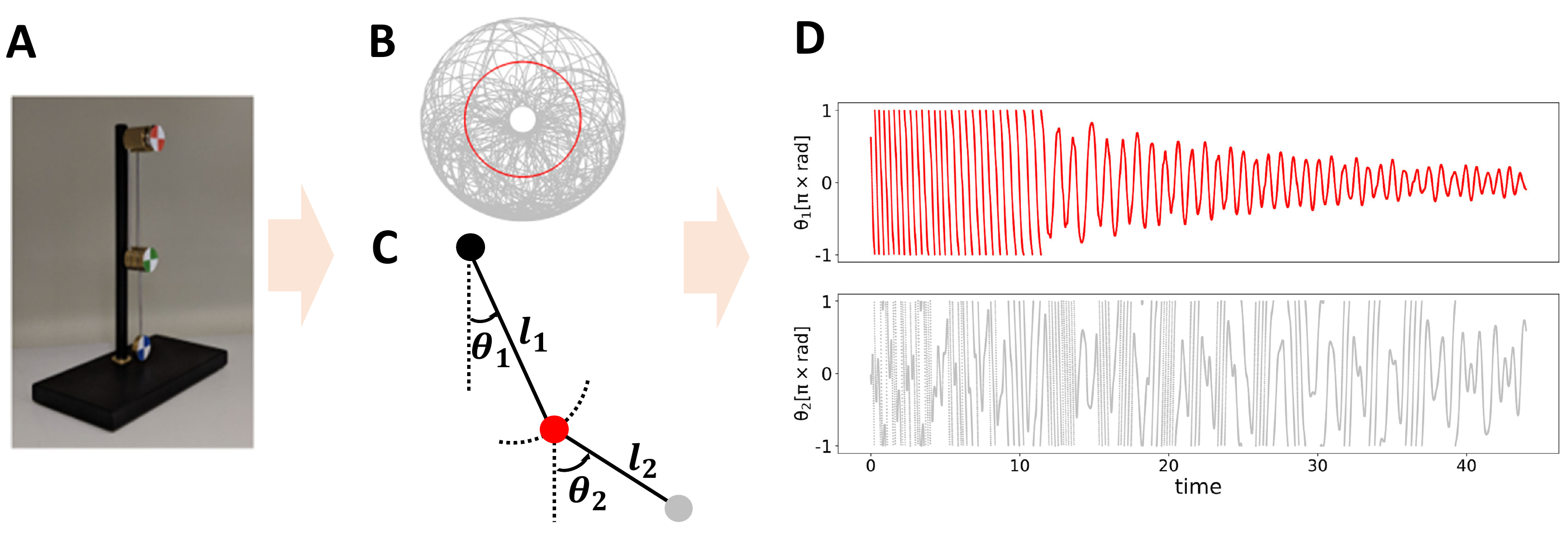} 
	\caption{Double Pendulum system experiment and measurement data \citep{asseman2018learning}: A. experiment setup. B. displacements of the two moving masses. C. model the system with $\theta_1$ and $\theta_2$. D. angles of two masses transformed from displacements. } 
	\label{Figure:DP}
\end{figure}

The second nonlinear dynamics discovery experiment is a chaotic double pendulum system \citep{asseman2018learning}. The \hsedit{measured} data, in form of videos, \hsedit{represents} the chaotic motion of a double pendulum on the device shown in Figure \ref{Figure:DP}A filmed with a high-speed camera. The positional data is converted into angular form \hsedit{based on the geometry information (see} the model shown in Figure \ref{Figure:DP}C). The governing equations can be derived \hsedit{using the} Euler–Lagrange method, \hsedit{given by}
\begin{equation} \label{eq:DQ1} 
\begin{split}
    (m_1+m_2)l_1 \ddot{\theta}_1 + m_2 l_2 \ddot{\theta}_2 \cos(\theta_1 - \theta_2) + m_2 l_2 \omega_2^2 \sin(\theta_1-\theta_2) + (m_1+m_2)g\sin(\theta_1) &= {\color{black} F_1}, \\
    m_2 l_2 \ddot{\theta}_2 + m_2 l_1 \ddot{\theta}_1 \cos(\theta_1 - \theta_2) -m_2 l_1 \omega_1^2 \sin(\theta_1 - \theta_2) + m_2 g \sin(\theta_2) &= {\color{black} F_2},
\end{split}
\end{equation}
\noindent which, by denoting $\omega$ as the velocity, can be converted to the \hsedit{following state-space} form:
\begin{equation} \label{eq:DQ2} 
\begin{split}
    \dot{\theta}_1&=\omega_1, \\
    \dot{\theta}_2&=\omega_2, \\
    \dot{\omega}_1&=c_{1}\dot{\omega}_2\cos(\Delta\theta)+c_{2}\omega_2^2\sin(\Delta\theta) + c_{3}\sin(\theta_1)+ \textcolor{black}{\mathcal{R}_1(\theta_1, \theta_2, \dot{\theta}_1, \dot{\theta}_2)}, \\
    \dot{\omega}_2&=c_{1}\dot{\omega}_1\cos(\Delta\theta)+c_{2}\omega_1^2\sin(\Delta\theta)+c_{3}\sin(\theta_2)+\textcolor{black}{\mathcal{R}_2(\theta_1, \theta_2, \dot{\theta}_1, \dot{\theta}_2)}
\end{split}
\end{equation}
where $\Delta\theta=\theta_1 - \theta_2$, and $\textcolor{black}{\mathcal{R}_1(\theta_1, \theta_2, \dot{\theta}_1, \dot{\theta}_2)}$ and $\textcolor{black}{\mathcal{R}_2(\theta_1, \theta_2, \dot{\theta}_1, \dot{\theta}_2)}$ denote the damping terms for the last two differential equations.

The data source contains multiple \hsedit{video} datasets. For this discovery, 5,000 denoised random sub-samples from 5 datasets are used for training purposes, and 2,000 random sub-samples from another 2 datasets for validation, and 1 dataset for testing. Some prior knowledge guiding this discovery includes:
\begin{enumerate}
    \item \hsedit{the two terms} $\dot{\omega}_2\cos(\Delta\theta)$ for $\dot{\omega}_1$ and $\dot{\omega}_1\cos(\Delta\theta)$ for $\dot{\omega}_2$, \hsedit{which can be easily derived based on our prior knowledge on the system, are assumed known in the two governing equations. However, their coefficients are unknown and need to be estimated}.
    \item the remaining of the formulas are potentially comprised of the free combination of $\dot{\omega}_1$, $\dot{\omega}_2$, $\omega_1$, $\omega_2$, as well as the angles ($\theta_1$, $\theta_2$, $\Delta\theta$) under the trigonometric functions $\cos(\cdot)$ and $\sin(\cdot)$.
    \item velocities and relative velocities of two masses, as well as their directions (sign function) might contribute to the damping.
\end{enumerate}
Based on \hsedit{the above} information, the candidate production rules for $\dot{\omega}_1$ and $\dot{\omega}_2$ \hsedit{equations} are shown below, where non-terminal nodes are $V=\{A,W,T,S\}$ and $C$ denotes the placeholder symbol for \hsedit{the} constant \hsedit{coefficient} values.
\begin{itemize}
    \item[] $A\rightarrow A+A$, $A\rightarrow A\times A$, $A\rightarrow C$, $A\rightarrow A+A$, $A\rightarrow W$,
    \item[] $W\rightarrow W\times W$, $W\rightarrow \omega_1$, $W\rightarrow \omega_2$, $W\rightarrow \dot{\omega}_1$, $W\rightarrow \dot{\omega}_2$,
    \item[] $A\rightarrow \cos(T)$, $A\rightarrow \sin(T)$, $T\rightarrow T+T$, $T\rightarrow T-T$, $T\rightarrow \theta_1$, $T\rightarrow \theta_2$,  
    \item[] $A\rightarrow sign(S)$, $S\rightarrow S+S$, $S\rightarrow S-S$, $S\rightarrow \omega_1$, $S\rightarrow \omega_2$, $S\rightarrow \dot{\omega}_1$, $S\rightarrow \dot{\omega}_2$,
    \item[] $A\rightarrow \dot{\omega}_1\cos(\theta_1-\theta_2)$, $A\rightarrow \dot{\omega}_2\cos(\theta_1-\theta_2)$.
\end{itemize}

\begin{figure}[t!]
	\centering
	\includegraphics[width=0.92\linewidth]{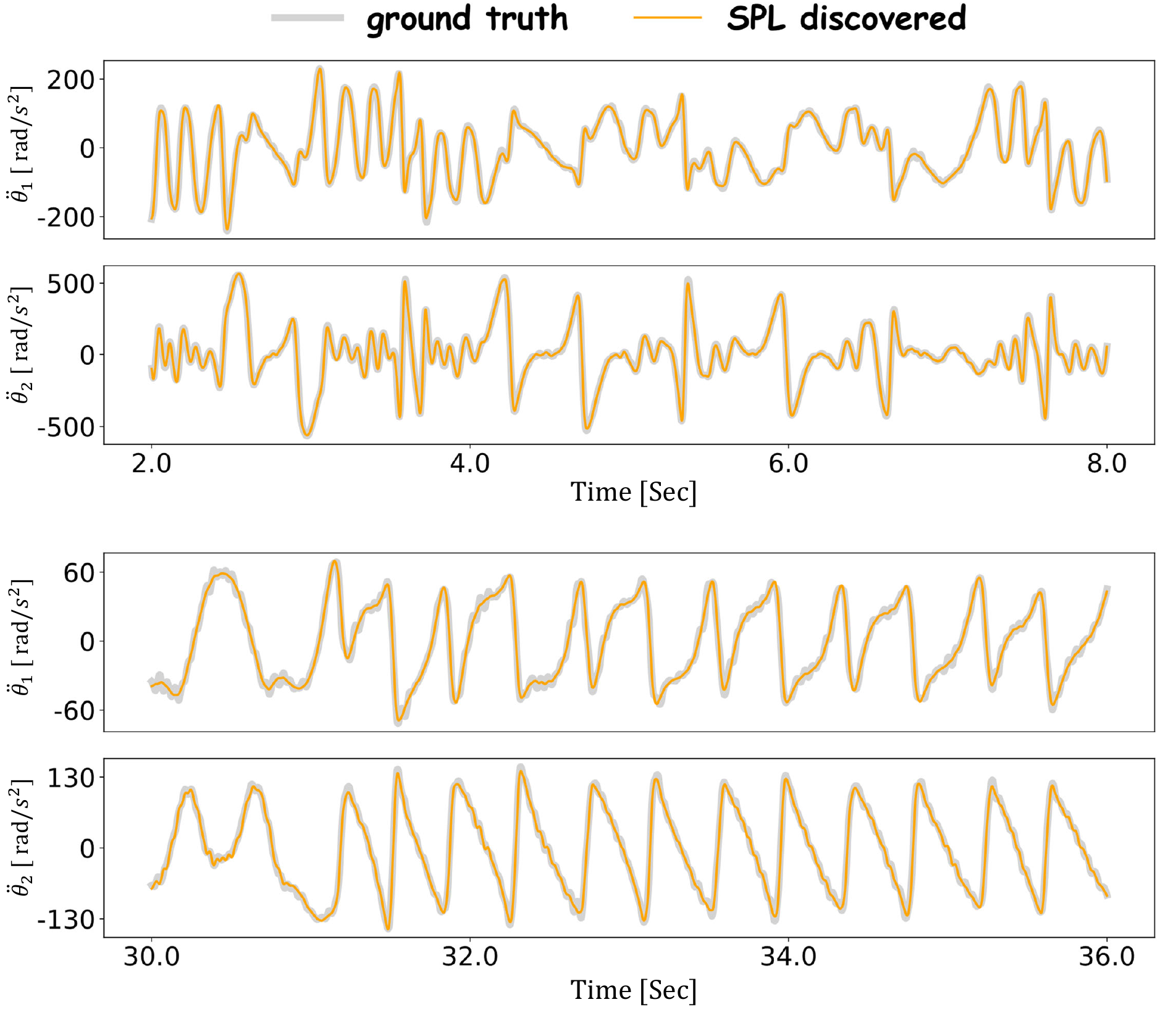}
	\caption{Discovered governing equations of the mounted double pendulum system on a different dataset in different time sections: in 2-8 seconds the two masses are in chaotic motions while in 30-36 seconds the masses tend to move periodically due to accumulative damping. $\ddot{\theta}_1$ and $\ddot{\theta}_2$ are obtained through smoothed numerical differentiation and predicted from the discovered governing equations.} 
	\label{Figure:s:dp_pred}
\end{figure}

\hsedit{Note that our prior knowledge can be easily incorporated in the proposed SPL machine to improve the discovery performance, rather than relying on \hsedit{the} free combination of mathematical operators and symbols.} The hyperparameters are set as $\eta=1$, $t_{max}=20$, and 40,000 episodes of training \hsedit{are} regarded as one trail. 5 independent trials are performed and the equations with the highest validation scores are selected as the final \hsedit{result}. The uncovered equations are shown in Table \ref{Table:s:dp}. They are validated through interpolation on the testing set and compared with the smoothed derivatives, as shown in Figure \ref{Figure:s:dp_pred}. The solution appears felicitous as the governing equations of the testing responses.

\begin{table}[t!]
    \centering
    \caption{Discovered governing equations of the mounted double pendulum by the SPL model. }
    \vspace{6pt}
    \begin{tabular}{cc}
    \toprule
    \textbf{Phase} &  \textbf{Expression} \\
    \midrule
    $\dot{\omega}_1$ & $-0.0991\dot{\omega}_2\cos(\Delta\theta)-0.103\omega_2^2\sin(\Delta\theta)-69.274\sin(\theta_1)+ \textcolor{black}{0.515\cos(\theta_1)}$ \\
    $\dot{\omega}_2$ & $-1.368\dot{\omega}_1\cos(\Delta\theta)+1.363\omega_1^2\sin(\Delta\theta)-92.913\sin(\theta_2)+\textcolor{black}{ 0.032\omega_1}$\\
    \bottomrule
    \end{tabular}
    \label{Table:s:dp}
    \vspace{0pt}
\end{table}

\begin{table}[t!]
\centering
\caption{Average training time (in seconds) of SPL and NGGP in the Nguyen's benchmark problems}
\vspace{6pt}
{
\begin{tabular}{lll}
\toprule
\textbf{Benchmark}  &  \textbf{SPL} [s] &  \textbf{NGGP} [s] \\
    \midrule
    Nguyen-1 & 8.776 & 2.734\\
    Nguyen-2 & 7.296 & 3.296\\
    Nguyen-3 & 81.287 & 3.945\\
    Nguyen-4 & 567.061 & 5.764\\
    Nguyen-5 & 431.228 & 77.627\\
    Nguyen-6 & 64.651 & 104.588\\
    Nguyen-7 & 14.995 & 3.024\\
    Nguyen-8 & 5.59 &  2.896\\
    Nguyen-9 & 5.743 & 13.229\\
    Nguyen-10 & 53.245 & 86.497\\
    Nguyen-11 & 10.163 & 44.399\\
    Nguyen-12 & 187.9 &  334.757\\
    Nguyen-1$^c$ & 452.734 & 362.075\\
    Nguyen-2$^c$ & 295.769 & 1188.215\\
    Nguyen-5$^c$ & 2178.891 & 1365.777\\
    Nguyen-8$^c$ & 77.892 & 129.349\\
    Nguyen-9$^c$ & 2001.402 & 3066.41\\
    \bottomrule
\end{tabular}
}
\label{Table:Nguyen Time}
\vspace{12pt}
\end{table}

\section{Discussion and Future Directions}\label{appendix:discussion}
\hsedit{The effectiveness of the proposed SPL machine is empowered by the following elements: (1) The use of MCTS enables the flexible representation of search space with customized computational grammars, composed of a finite set of mathematical operators and symbols, to guide the search tree expansion. (2) The exploration-exploitation trade-off nature of MCTS is remarkably useful for searching the optimal mathematical expression tree. (3) The key adjustments, including the greedy search, the adaptive-scaled rewarding, the reward regularizer, and the expression tree module transplantation, make it possible to efficiently uncover the best path to formulate complex equations. (4) The SPL machine straightforwardly accepts our prior or domain knowledge, or any sort of constraints of the tasks in the grammar design while leveraging great flexibility in expression formulation.}

While SPL shows huge potential in both symbolic regression and governing equation discovery tasks, there are still some imperfections to be improved. In this appendix section, a few bottlenecks and their potential solutions are presented:
\begin{enumerate}
    \item \textbf{Computational cost.} Computational cost for this framework is one of the major issues, especially when constant \hsedit{coefficient} estimation is required. Evaluating the solution in the simulation phase happens very frequently for the MCTS algorithm where the policy selection relies heavily on a large number of historical rewards. However, the constant \hsedit{coefficient} value estimation, which \hsedit{requires repeated calls for an optimization process} and can be slow, is \hsedit{needed} for evaluation purposes. \hsedit{In particular, the constant coefficient value is estimated via concurrently solving an optimization problem, e.g., by Powell's conjugate direction method \citep{powell1964efficient}. For example, when the tree structure changes or is updated, the optimization of the constant coefficients should be re-performed simultaneously.} The SPL machine is not the only symbolic regressor suffering from the computational cost in constant \hsedit{coefficient} value estimation. In fact, the state-of-the-art symbolic regression model, the neural-guided GP (NGGP) \citep{mundhenk2021symbolic}, becomes much slower in the Nguyen's benchmark variant tasks (see Table \ref{Table:Nguyen Time}). The current implementation of the SPL machine tries to empirically avoid this issue by limiting the number of placeholders in a discovered expression and simplifying the expression before \hsedit{evaluation}, but still cannot reach \hsedit{great} efficiency. This bottleneck might be mitigated if parallel computing is introduced to the MCTS simulation phase.
    \item \textbf{Graph modularity underexamined.} The current design of the SPL training scheme does not leverage the full graph modularity: modules are reached by transforming a complete parse tree into a grammar. However, in some cases, there might be some influential modules appearing frequently as part of the tree. This type of graph modularity is described in the AI-Feynman method \citep{udrescu2020ai2}. Deploying a more comprehensive graph modularity into the SPL machine will boost its efficacy in \hsedit{the} complex equation and nonlinear dynamics discovery tasks. 
    \item \textbf{Robustness against extreme data noise and scarcity.} \hsedit{Although} it is observed that this reinforcement learning-based method is able to unearth the parsimonious solution to the governing equations from synthetic or measurement data with a moderate level of noise and scarcity. It is not effective when the data condition is extreme, or if there are missing values that make it challenging to numerically calculate the state derivatives. It is reasonable to investigate the integration between the SPL framework with a \hsedit{differentiable} surrogate model built upon neural networks \citep{long2018pde,chen2021physics} or spline learning \citep{sun2021physics} for further robustness in nonlinear dynamics discovery tasks.
\end{enumerate}

\end{document}